\documentclass[fleqn,10pt]{wlscirep}
\usepackage[utf8]{inputenc}
\usepackage[T1]{fontenc}
\usepackage{framed}
\usepackage{multirow}
\usepackage{tabulary}
\usepackage{tabularx}
\usepackage{amssymb}
\usepackage{bbold}
\usepackage{latexsym}
\usepackage{comment}
\usepackage{color, colortbl}
\usepackage{url}
\usepackage{xcolor}
\usepackage[super]{nth}
\usepackage{makecell}
\usepackage{float}
\usepackage[utf8]{inputenc}
\usepackage[english]{babel}
\usepackage{bigdelim}
\usepackage{graphicx}
\usepackage{enumitem}
\usepackage{amsmath}
\usepackage{hyperref}
\usepackage{multicol}
\usepackage{amsfonts}
\usepackage{subfig}
\usepackage{pifont}
\usepackage[normalem]{ulem}

\definecolor{Gray}{gray}{0.9}
\definecolor{newcolor}{rgb}{.8,.349,.1}
\newcommand\blfootnote[1]{%
  \begingroup
  \renewcommand\thefootnote{}\footnote{#1}%
  \addtocounter{footnote}{-1}%
  \endgroup
}

\newcommand{\cmark}{\ding{51}}
\newcommand{\xmark}{\ding{55}}

\hypersetup{
    colorlinks=true,
    linkcolor=green,
    filecolor=magenta,      
    urlcolor=blue,
}

\title{Benchmarking CNN on 3D Anatomical Brain MRI: Architectures, Data Augmentation and Deep Ensemble Learning}

\author[1,2,*]{Benoit Dufumier}
\author[2]{Pietro Gori}
\author[2]{Ilaria Battaglia}
\author[1]{Julie Victor}
\author[1]{Antoine Grigis}
\author[1]{Edouard Duchesnay}

\affil[1]{NeuroSpin, CEA Saclay, Université Paris-Saclay, France }
\affil[2]{LTCI, Télécom Paris, IPParis, France}

\affil[*]{benoit.dufumier@cea.fr}


\begin{abstract}
Deep Learning (DL) and specifically CNN models have become a \textit{de facto} method for a wide range of vision tasks, outperforming traditional machine learning (ML) methods. Consequently, they drew a lot of attention in the neuroimaging field in particular for phenotype prediction or computer-aided diagnosis. However, most of the current studies often deal with small single-site cohorts, along with a specific pre-processing pipeline and custom CNN architectures, which make them difficult to compare to.
We propose an extensive benchmark of recent state-of-the-art (SOTA) 3D CNN, evaluating also the benefits of data augmentation and deep ensemble learning, on both Voxel-Based Morphometry (VBM)  pre-processing and minimally pre-processed \textit{quasi-raw} images. Experiments were conducted on a large heterogeneous multi-site 3D brain anatomical MRI data-set comprising $N=10$k scans on 3 challenging tasks: age prediction, sex classification, and schizophrenia diagnosis.
We found that all models provide significantly better predictions with VBM images than quasi-raw data. This finding evolved as the training set approaches 10k samples where quasi-raw data almost reach the performance of VBM. Moreover, we showed that linear models perform comparably with SOTA CNN on VBM data. We also demonstrated that DenseNet and tiny-DenseNet, a lighter version that we proposed, provide a good compromise in terms of performance in all data regime. Therefore, we suggest to employ them as the architectures by default.
Critically, we also showed that current CNN are still very biased towards the acquisition site, even when trained with $N=10$k multi-site images. In this context, VBM pre-processing provides an efficient way to limit this site effect.
Surprisingly, we did not find any clear benefit from data augmentation techniques - and more recently proposed MRI artefacts for brain MRI.
Finally, we also showed that big CNN models were not well calibrated when trained with small brain MRI data-sets and we empirically proved that deep ensemble learning is well suited to re-calibrate them without sacrificing performance. 
\end{abstract}

\begin{document}

\flushbottom
\maketitle

\thispagestyle{empty}

\blfootnote{Technical report}
\section{Introduction}


    Since the breakthrough in 2012 of AlexNet \cite{alexnet2012} during the ILSVRC-2012 challenge, Convolutional Neural Networks (CNN) gained a lot of attention in the computer vision community. In the following years, they proved to be the SOTA for various computer vision tasks where enough data were available (typically $N>10^6$); among them, object detection \cite{girshick2015rcnn}, semantic segmentation \cite{badrinarayanan2017segnet}, image denoising \cite{zhang2017beyond}, etc. Several architectures \cite{VGG_Simonyan, ResNet_He, ResNeXt_Xie, szegedy2016inception_v3, DenseNet_Huang} have been proposed over the years to constantly improve the performance of the networks in particular for 2D image classification on natural images (\textit{e.g} CIFAR \cite{krizhevsky2009learning}, ImageNet \cite{deng2009imagenet}, MNIST \cite{lecun1998gradient}). A key advantage of CNN-based models is that they do not require manual extraction of hand-crafted features and they are able to learn high-level abstractions of images in a hierarchical manner by using back-propagation. However, one main drawback is their need for massive amount of data to converge properly. \\
    
    As a result, they drew a lot of attention in the neuroimaging field as large open-access MRI databases were becoming available (\textit{e.g} UKBiobank \cite{bycroft2018ukb} or
    the Human Connectome Project \cite{van2013hcp}).
    Deep Neural Networks (DNNs) have been used in numerous neuroimaging applications such as image registration \cite{yang2017quicksilver}, tumor detection \cite{havaei2017braintumor}, or brain disease prediction (\textit{e.g.} Alzheimer's detection \cite{wen2020}, schizophrenia \cite{plis2014deep} or autism \cite{sujit2019automated, shahamat2020brain}). \\
    
    Different studies, focusing on 3D neuroanatomical MRI data, proposed custom CNN architectures based on recent advances in computer vision to perform various regression or classification tasks (see table \ref{comparison_neuroimage_cnn} for a detailed review). However, none of these papers compared their performance with other SOTA CNN networks. Furthermore, most of them used different pre-processing pipelines and datasets with various size ($N$ ranging from several hundreds to several thousands, see table \ref{comparison_neuroimage_cnn}), making it difficult to compare them. Even if some benchmarks started to emerge such as \cite{wen2020} for Alzheimer's detection with anatomical MRI or \cite{schulz2020, Jonsson2019} for phenotype prediction, they are still difficult to reproduce because of a costly pre-processing pipeline and they still lack a fair comparison between SOTA CNN models.\\
    
    Furthermore, over-fitting is quite common with currently available MRI datasets since they are relatively small compared to standard natural images ones ($N<10^4$ vs $N>10^6$). Data augmentation is one way of limiting this effect by adding artificial samples to the training set and it was employed in most papers (see table \ref{comparison_neuroimage_cnn}). These samples are generated by deforming the images in the training set while preserving its semantic information for the target prediction task (\textit{e.g} by cropping, translating or rotating the images). Nonetheless, there is currently no consensus on the applicability of these transformations to MRI data. \\
    
    Finally, modern CNN models are known to be over-confident in their prediction, especially for classification tasks \cite{guo2017calibration}. The quantification of their epistemic uncertainty \cite{gal2016phd} (inherent to the model) is thus of primary importance for clinical applications. Here, we aim at comparing the epistemic uncertainties of SOTA CNNs in the small data regime($N < 10^3$), which is the typical size when dealing with clinical cohorts. Two main scalable methods were proposed in the literature to tackle it: Deep Ensemble learning \cite{lakshminarayanan2017} and MC-Dropout \cite{gal2016phd}. Deep Ensemble learning has several advantages compared to MC-Dropout: it does not modify the CNN architecture, it is very easy to implement, it generally leads to better performance during challenges (\textit{e.g.} AlexNet, VGG or GoogLeNet for ILVSRC) and it needs very little hyperparameter tuning \cite{lakshminarayanan2017}. In addition, a recent study \cite{gustafsson2020bench} has shown that Deep Ensemble learning consistently gives better uncertainty estimates for real-world classification and regression tasks on natural images. It is thus particularly suited for this study\footnote{We also performed experiments with MC-Dropout. The details can be found in the Supplementary and the results are discussed in section \ref{discussion_sec}.}. 
    
    \subsection{Contributions}
    In this paper, we propose to benchmark SOTA CNN architectures on a large-scale multi-centric brain MRI dataset comprising $N=10$K scans of healthy participants, namely BHB-10K, pre-processed with two different pipelines: minimally prepocessed \textit{quasi-raw} data and Voxel-Based Morphometry (VBM) \cite{ashburner2007dartel}, see section~\ref{preproc_details}). We also aim at giving the first benchmark of CNN models for schizophrenia's prediction using two independent clinical datasets (see table \ref{table:demographic_infos}). We show that the pre-processing, as well as the data regime, are critical when performing DL with neuroimaging data. Specifically, we demonstrate that CNN perform equally well in the low data regime $N\le 10^3$), no matter the depth or architecture, and that linear models are still competitive, given an appropriate extensive pre-processing. Secondly, in the big data regime $N=10^4$, CNN perform better than linear models, no matter the pre-processing, but given a sufficiently deep architecture. Critically, we also demonstrate that all models are currently very biased towards the site and that extensive non-linear pre-processing provides a simple way to limit this effect. We show that data augmentation brings little or no improvement in the small data regime ($N$=500). Finally, we demonstrate that big CNN models are mis-calibrated in this regime but  deep ensemble learning provides an efficient way to re-calibrate them while even improving the performance.\\
    
    As a step towards reproducible research, we provide an open access to the Python code and to the BHB-10K dataset, pre-processed with our 2 different pipelines (quasi-raw and VBM) here\footnote{Some data-sets still may not be released because of authorization issues but the releasing process is on-going.}:\\
    \url{https://github.com/Duplums/bhb10k-dl-benchmark} \\
    
    As opposed to UKBioBank or HCP, this dataset is highly multi-centric and the images have been acquired with various protocols and spatial resolutions. More details about the dataset can be found in section \ref{bhb10k_description}.
    
\section{Related Works}

        \begin{table*}[h!]
        \setlength\tabcolsep{2pt}
            \centering
            \begin{tabular}{|>{\centering}m{0.15\textwidth}|>{\centering}m{0.05\textwidth}|>{\centering}m{0.1\textwidth}|>{\centering}m{0.05\textwidth}|>{\centering}m{0.2\textwidth}|>{\centering}m{0.2\textwidth}|>{\centering\arraybackslash}m{0.1\textwidth}|}
                \hline
                \rowcolor{Gray}
                 \textbf{Task} & \textbf{Study} & \textbf{Backbone} & \textbf{N} & \textbf{Preprocessing} & \textbf{Data Augmentation} & \textbf{Ensemble Learning}  \\
                 \hline
                 \hline
                 \multirow{8}{*}{Age Prediction} & \cite{peng2021} & VGG & $14K$ & Quasi-Raw & Translation and Flip & \cmark \\
                                                 & \cite{sturmfels2018domain} & VGG & $724$ & VBM & \xmark & \xmark \\
                                                 & \cite{cole2017predicting} & VGG &$2001$ & Quasi-Raw and VBM & Translation and Rotation & \xmark\\
                                                 & \cite{ueda2019age} & VGG & $1101$ & VBM & Translation, Crop and Flip& \xmark \\
                                                 & \cite{Jonsson2019} & ResNet & $1264$ & Quasi-Raw and VBM & Translation and Rotation & \cmark \\
                                                 & \cite{armanious2020age} & Inception and SqueezeNet & $562$ & VBM & Translation and Flip & \xmark \\
                                                & \cite{varatharajah2018} & Inception & $12988$ & Raw & \xmark & \xmark \\
                                                & \cite{bashyam2020mri} & Inception & 11729 & Raw & Flip and Intensity Scaling & \xmark \\
                \hline
                 \multirow{10}{*}{AD vs HC} & \cite{li2017ad} & VGG & $427$ & VBM & \xmark & \xmark \\
                                            & \cite{wen2020} & VGG & $1455$ & Quasi-Raw and VBM & \xmark & \xmark\\
                                            & \cite{backstrom2018efficient} & VGG & $1198$ & Quasi-Raw & Flip & \xmark \\
                                            & \cite{korolev2017} & VGG and ResNet & $231$ & Quasi-Raw & \xmark & \xmark \\
                                            & \cite{hosseini2018} & VGG & $210$ & Unclear & \xmark & \xmark \\
                                            & \cite{shmulev2018} & VGG and ResNet & $2780$ & Quasi-Raw & \xmark & \xmark \\
                                            & \cite{abrol2020} & ResNet & $828$ & VBM & \xmark & \xmark \\
                                            & \cite{senanayake2018} & ResNet and DenseNet & $515$ & Unclear & \xmark & \xmark \\
                                            & \cite{spasov2019} & ResNet & $785$ & VBM & \xmark & \xmark \\
                                            & \cite{wang2019} & DenseNet & $833$ & Quasi-Raw & \xmark & \cmark \\
                \hline
                \multirow{2}{*}{ASD vs CTL} & \cite{sujit2019automated} & VGG & $1064$ & Quasi-Raw & Zooming, Affine and Flip & \cmark  \\
                                            & \cite{shahamat2020brain} & VGG & $935$ & Quasi-Raw & \xmark & \xmark \\
                \hline
                \multirow{2}{*}{SCZ vs CTL} & \cite{hu2020} & VGG, ResNet and Inception & $450$ & VBM & \xmark & \xmark \\
                                            & \cite{oh2020} & VGG & $866$ & Quasi-Raw & \xmark & \xmark \\
                \hline
                \multirow{1}{*}{Sex Classification} & \cite{peng2021} & VGG & $6K$ & Quasi-Raw & Translation and Flip & \cmark \\
                \hline
            \end{tabular}
            \caption{Summary of studies tackling 5 clinical problems using 3D neuroanatomical MRI data with various kind of SOTA 3D CNN backbones. AD: Alzheimer's Disease; ASD: Autism Spectrum Disorder; SCZ: schizophrenia; HC: Healthy Control; N: total number of samples in the dataset; Quasi-Raw: minimal pre-processing including mostly linear registration to the MNI template; VBM: non-linear registration to the MNI template along with segmentation of the tissues and brain extraction.  Note: for AD vs HC, we did not report the study with clear data leakage as defined in \cite{wen2020}. }
            \label{comparison_neuroimage_cnn}
    \end{table*}

    Very recent studies tackled brain age prediction with CNN using 3D neuroanatomical MR images \cite{cole2017predicting, peng2021, sturmfels2018domain, ueda2019age, Jonsson2019, armanious2020age, bashyam2020mri}. Most of these works used a single pre-processing (either VBM or Quasi-Raw, cf. table \ref{comparison_neuroimage_cnn}) and the classical VGG backbone architecture (repetition of blocks \textit{Convolution-Batch Normalization-ReLu} with a \textit{MaxPooling} layer between each block and a kernel size $3\times3\times3$ for the convolutional layers). Notably, \cite{armanious2020age} used inception modules followed by a fire module inspired by Inception v3 \cite{szegedy2016inception_v3} and SqueezeNet \cite{iandola2016squeezenet} while \cite{Jonsson2019} used the classical ResNet architecture. \cite{bashyam2020mri} is the first paper to use a large-scale Inception-based network for transfer learning with age prediction as pre-training. 
    
    Classification between schizophrenic patients and Healthy Controls (HC) based on neuroanatomical differences, \textit{e.g} cortical thinning in prefontal and temporal regions and volume reductions in thalamus, has been widely studied with traditional ML methods such as Support Vector Machine (SVM) \cite{gould2014svm_scz, depierrefeu2018identifying, xiao2019svm, vieira2020using}, Deep Belief Network \cite{plis2014deep, pinaya2016dbn, latha2019} or Stack Auto-Encoder \cite{pinaya2019} with shallow neural network architectures. Recent studies \cite{oh2020, hu2020} are starting to tackle this task without a feature extraction step with a DL approach but they are still limited to custom 3D architectures (\textit{e.g} designed for video classification in \cite{oh2020}).
    
    Very few works proposed a benchmark between 3D CNN architectures. Among them, in \cite{hu2020}, authors compared VGG, ResNet and Inception to discriminate between schizophrenic patients and control subjects, but they limited their tests to shallow architectures with few layers and only to data pre-processed with VBM.
    
    In \cite{wen2020}, authors proposed to benchmark 2 pre-processing pipelines (named Minimal and Extensive) along with different input dimensions (3D subject-level, 2D slice-level, Region-Of-Interest or 3D patch-level) and Transfer Learning strategies for Alzheimer's disease classification with anatomical MRI. They found no difference between 3D-ROI, 3D subject-level and 3D patch-level approach while the 2D slice-level approach performed worse. 
    However, all tested architectures shared the VGG backbone and they did not integrate other SOTA improvements such as skip-connection (ResNet), inception module (Inception) or feature re-using (DenseNet). 
    
    Authors in \cite{Jonsson2019} compared classical ML algorithms (Ridge Regression, Gaussian Process Regression) with a ResNet-based CNN architecture on age prediction. They found that CNN model performs better than ML methods when trained on several brain tissues and Jacobian map extracted from T1-weighted images. They also observed a site effect when they tested their algorithm on an independent data-set (+70\% $l_1$ error when tested on IXI dataset  without transfer learning).
    
    Last year, the authors of \cite{schulz2020} systematically compared DNN models (both MLP and CNN) with linear models and non-linear SVM on age and sex classification using  anatomical MRI, as the number of training samples increases. They conclude that DNN models perform equally well than traditional ML algorithms. Differently from our study i) they treated age and sex classification together by performing a 10-class classification, ii) they used feature selection to perform the classification, iii) their purpose was not to compare CNN models but rather to compare DNN with ML models, iv) their study used solely UKBiobank to train and test their model (only one acquisition protocols with 3 identical scanners) while we propose a new benchmark on a highly multi-centric brain MRI dataset pre-processed with 2 pipelines, and openly accessible.
    
    More recently, \cite{abrol2021deep} showed that DNN scale very well for age and sex prediction with  anatomical MRI, given a large homogeneous dataset (UKBioBank with $N=10^4$ in their case). As opposed to \cite{schulz2020}, they critically demonstrated that DNN can provide a relevant representation for the task at hand (better than most ML approaches), when no feature extraction step is performed beforehand. However, their work is also limited to a big homogeneous dataset and a comprehensible study of CNN architectures is still lacking when dealing with MR images.

\section{Material and Methods}

    Here, we present the datasets used throughout the experiments for age and sex prediction on healthy cohorts (regression and classification tasks respectively), in section \ref{bhb10k_description}, and schizophrenia's prediction (Dx), in section \ref{clinical_datasets_description}. Furthermore, even if CNN are known to perform well on raw data \cite{goodfellow2016deep, lecun2015deep} (at least on vision tasks, \textit{e.g.} classification with ImageNet \cite{deng2009imagenet}), it is still not clear whether their performance on neuroimaging data can be impacted by an extensive preprocessing and to what extent it depends on the training size. To answer these important questions and similarly to \cite{cole2017predicting} and \cite{wen2020}, we studied 2 kinds of preprocessing, namely \textbf{VBM} and \textbf{Quasi-Raw} detailed in section \ref{preproc_details}. The architectures of SOTA CNN are described in section \ref{cnn_archi_details} and the data augmentation and deep ensemble strategies can be found in sections \ref{data_augmentation_strats} and \ref{deep_ensemble} respectively.

    \subsection{BHB-10K Dataset}
    \label{bhb10k_description}
        
        We aggregated 13 publicly available data-sets coming from various data-sharing initiatives and including $N=10420$ T1-weighted 3D MRI scans for 7764 healthy individuals (with several sessions per participant in some cases). The acquisitions were performed with either 1.5T or 3T scanners with potentially different acquisition protocols across sites (see table \ref{table:demographic_infos} and the data-set sources for more details).
    
    \subsection{Clinical Datasets}
        \label{clinical_datasets_description}
        
        In addition to BHB-10K, we also gathered 2 other independent multi-site data-sets, namely SCHIZCONNECT-VIP\footnote{http://schizconnect.org} and Bipolar and Schizophrenia Network for Intermediate Phenotype (BSNIP) \cite{tamminga2014bipolar}, including both healthy controls and patients with strict schizophrenia and also composed of T1-weighted 3D MRI scans (see table \ref{table:demographic_infos}). SCHIZCONNECT-VIP combines 4 publicly available cohorts of controls and patients with schizophrenia. These cohorts are heterogeneous both in terms of acquisition scanners and geographical sites. As for BSNIP, the MR images were acquired on 5 different centers with 3T scanners spread across the USA. It contains cohorts of healthy controls and patients with schizophrenia and all the clinical assessments were standardized. Crucially, BSNIP is only used as a test set throughout this study while SCHIZCONNECT-VIP and BHB-10K are used for training and validation (see table \ref{training_test_split_supp} in the Supplementary)
        
        \begin{table}[h!]
            \centering
            \resizebox{\textwidth}{!}{  
            \begin{tabular}{|l c|c|c|c|c|c|c|c|c}
                \hline
                & \textbf{Datasets} & \textbf{Diagnosis} & \textbf{\# Subjects} & \textbf{N} & \textbf{Age} & \textbf{Sex (\%F)} & \textbf{\# Sites} \\
                \hline
                & \multirow{2}{*}{\href{(http://schizconnect.org}{SCHIZCONNECT-VIP}} & schizophrenia & 275 & 275 & $34 \pm 12$ & 27 & 4\\
                                                   & & control & 330 & 330 & $32 \pm 12$ & 47 & 4 \\
                \hline
                & \multirow{2}{*}{BSNIP \cite{tamminga2014bipolar}} & schizophrenia & 194 & 194 & $34\pm 12$ & 44 & 5 \\
                                       & & control & 200 & 200 & $38\pm 13$ & 58 & 5 \\
                \hline
                & BIOBD \cite{hozer2020biobd} & control & 356 & 356 & $40\pm 13$ & 55 & 8 \\
                \hline
                \ldelim\{{13}{5mm}[\href{https://github.com/Duplums/bhb10k-dl-benchmark}{BHB-10K}] & \href{https://www.humanconnectome.org/study/hcp-young-adult}{HCP} & \multirow{13}{*}{control} & 1113 & 1113 & $29 \pm 4$ & 45 & 1 \\
                & \href{http://brain-development.org/ixi-dataset}{IXI} & & 559 & 559 & $48 \pm 16$ & 55 & 3\\
                & \href{https://www.nitrc.org/projects/fcon_1000/}{CoRR} &  & 1371 & 2897 & $26 \pm 16$ & 50 & 19\\
                & \href{https://openneuro.org/datasets/ds002330/versions/1.1.0}{NPC} &  & 65 & 65 & $26 \pm 4$ & 55 & 1 \\
                & \href{https://openneuro.org/datasets/ds002345/versions/1.0.1}{NAR} &  & 303 & 323 & $22 \pm 5$ & 58 & 1 \\
                & \href{https://openneuro.org/datasets/ds002247/versions/1.0.0}{RBP} &  & 40 & 40 & $22 \pm 5$ & 52 & 1 \\
                & \href{https://www.oasis-brains.org}{OASIS 3} &  & 597 & 1262 & $68\pm 9$ & 62 & 3 \\
                & \href{https://dataverse.harvard.edu/dataset.xhtml?persistentId=doi:10.7910/DVN/25833}{GSP} &  & 1570 & 1639 & $21 \pm 3$ & 58 & 1\\
                & \href{https://ida.loni.usc.edu}{ICBM} &  & 622 & 977 & $30 \pm 12$ & 45 & 3\\
                & \href{http://fcon_1000.projects.nitrc.org/indi/abide}{ABIDE I} &  & 567 & 567 & $17 \pm 8$ & 17 & 17\\
                & \href{http://fcon_1000.projects.nitrc.org/indi/abide}{ABIDE II} &  & 559 & 580 & $15 \pm 9$ & 30 & 19\\
                & \href{http://brainomics.cea.fr/localizer/localizer}{Localizer} &  & 82 & 82 & $25 \pm 7$ & 56 & 2\\
                & \href{https://openneuro.org/datasets/ds000221/versions/00002}{MPI-Leipzig} &  & 316 & 316 & $37 \pm 19$ & 40 & 2\\
                & \textbf{Total} & & 7764 & \textbf{10420} & $32\pm 19$ & 50 & 73 \\
                \hline
            \end{tabular}
             }
            \captionof{table}{Demographic information about the datasets. The number of sites indicates the number of acquisition MRI scanners used in the study. Each scanner does not necessarily use the same magnetic field intensity (e.g in the IXI study, three different scanners were used: a Philips 3T, a Philips 1.5T and a GE 1.5T).\label{table:demographic_infos}}
        \end{table}

    \subsection{Preprocessing}
        \label{preproc_details}
        
        \begin{figure*}[h!]
            \centering
            \includegraphics[width=.7\linewidth]{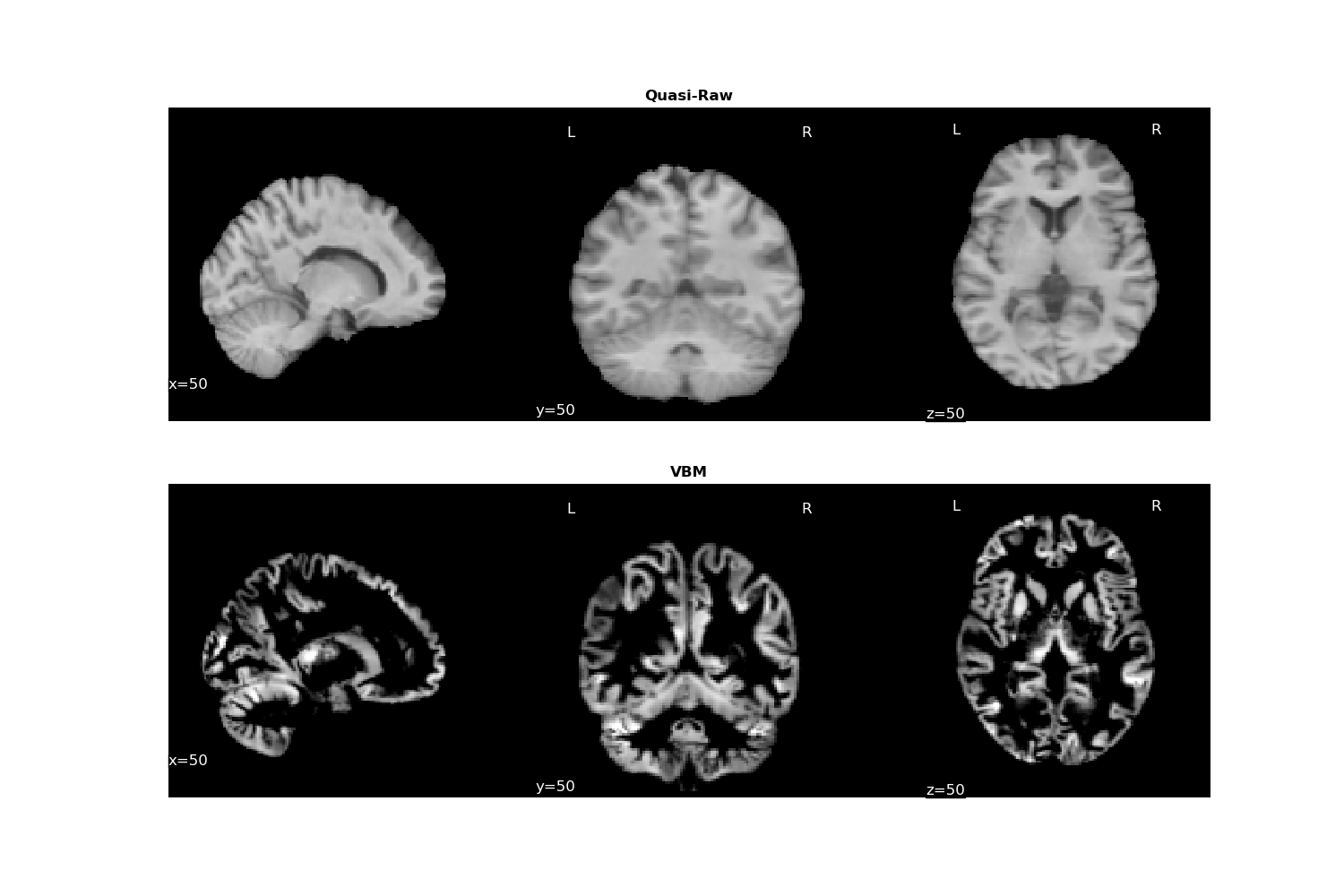}
            \caption{\textbf{Example of a 3D input image pre-processed with the Quasi-Raw framework (top) and the VBM procedure (bottom).} At the top, the image is resampled to $1.5mm^3$ isotropic, linearly registered to the MNI template and the brain is extracted. At the bottom, a non-linear registration algorithm is applied with the DARTEL algorithm \cite{ashburner2007dartel} and the image is modulated by the Jacobian of the deformation field. It is then resampled to $1.5mm^3$ isotropic.}
            \label{fig:img_example_quasi_raw_cat12}
        \end{figure*}
        
        \subsubsection{Voxel-Based Morphometry}
        
            The VBM pre-processing was performed with CAT12 \cite{gaser2016cat12vbm} from the SPM toolbox. It essentially consisted of a correction of the bias field and the noise in MRI images, the segmentation of Gray Matter (GM), White Matter (WM), and the cerebrospinal fluid (CSF). Images were normalized into a common standard MNI space composing a linear transformation that accounts for global alignment (rotation, translation, and global brain size) with a non-linear deformation \cite{ashburner2007dartel} that locally aligns brain structures. The normalized images are finally modulated by the Jacobian of their transformation to preserve the quantity of tissue. The images were re-sampled to an isotropic $1.5mm^3$ spatial resolution. The final output dimension is $121\times 145\times 121$. We retained only 2,122,945 voxels of GM maps. To remove the Total Intracranial Volume (TIV) co-variate effect, each GM map was normalized by the TIV estimated by CAT12 during the segmentation step. Note that this step cancels the part of the Jacobian that stems from the initial linear transformation. Finally, we also applied a visual quality check and we removed images poorly segmented or with obvious MR artefacts.
                
        \subsubsection{Quasi-raw data}
        
            This pre-processing was designed to be minimal. Consequently, only essential steps have been kept in order to map the images coming from different sites and scanners to the same space with the same resolution and only important image correction steps have been applied. Specifically, each scan is re-oriented to the MNI space and then re-sampled to a 1.5$mm^3$ spatial resolution through a linear spline interpolation. In the Supplementary (see table \ref{tab:spatial_res_impact}), we show that re-sampling at a higher resolution (\textit{e.g} 1$mm^3$) does not improve  the results. The bias field is corrected using the N4ITK algorithm \cite{tustison2010n4itk} from ANTs \cite{avants2009ants} and the brain is extracted with BET2 \cite{jenkinson2005bet2} (the skull and non-brain tissues are removed). Each image is linearly registered (9 degrees of freedom) to the MNI template with FLIRT from FSL \cite{jenkinson2001flirt}. Finally, we also applied a visual QC to the output images. An example of a scan pre-processed with these 2 pipelines is shown in figure \ref{fig:img_example_quasi_raw_cat12}.

    \subsection{CNN Models}
    \label{cnn_archi_details}
        
        In this work, we selected four CNN architectures which are widely used in computer vision and in most neuroimaging studies (see table \ref{comparison_neuroimage_cnn}). Specifically, we considered VGG \cite{VGG_Simonyan}, ResNet \cite{ResNet_He}, ResNeXt \cite{ResNeXt_Xie} (which combined the ideas from ResNet and Inception \cite{szegedy2016inception_v3}) and DenseNet \cite{DenseNet_Huang}. 
        
        \textbf{VGG} has a lot of different adaptations for different applications in the neuroimaging field \textit{e.g} Alzheimer's detection \cite{li2017ad, wen2020, backstrom2018efficient}, age prediction \cite{cole2017predicting, sturmfels2018domain}, etc. The core idea is to stack multiple layers, typically following the scheme \textit{Convolution-Batch Normalization-ReLu}, with a small kernel size, usually equal to $3$, for each convolution layer. Five blocks are usually used between each \textit{MaxPooling} layer and the number of channels inside each block is typically set to 64, 128, 256, 512 and 512 respectively. 
        
        \textbf{tiny-VGG} \cite{cole2017predicting} is currently the SOTA algorithm for age prediction on the BAHC \cite{cole2017predicting} dataset and it has been designed based on VGG11, with 8 times less channels per block and a small Fully-Connected layer at the end. It is the smallest CNN included in this study with 800K parameters. 
        
        \textbf{SFCN} \cite{peng2021} is another network designed for age and sex prediction (SOTA on the UKBiobank dataset \cite{bycroft2018ukb}) and it is also based on VGG. The main difference resides in a deeper architecture with 7 blocks, more channels per block, and a dropout layer put at the end in order to reduce over-fitting. In \cite{peng2021}, the authors introduced also a Data Augmentation strategy and they considered the age prediction problem as a classification problem. Here, as we want to give a fair comparison between SOTA CNN architectures, we did not use this strategy. 
        
        \textbf{ResNet} \cite{ResNet_He} introduced skip-connections to avoid the vanishing-gradient issue, often observed with very deep CNN architectures, and to prevent over-fitting. This allows the use of more layers and parameters without losing the generalization capacity of the models. It has been shown to perform well on UKBioBank and IXI \cite{Jonsson2019} for age prediction. Consequently, we compared 3 ResNet models with various depth and size (namely ResNet18, ResNet34 and ResNet50). 
        
        \textbf{ResNeXt} \cite{ResNeXt_Xie} integrates the advances from ResNet and Inception, making CNNs deeper and wider, while preserving the same number of parameters and FLOPs complexity of ResNet. We chose the ResNeXt50 model to have a direct comparison with ResNet50. 
        
        Finally, \textbf{DenseNet} \cite{DenseNet_Huang} introduced the concept of feature re-using. It is lighter than all the previous networks (except for tiny-VGG and SFCN) while it performs better than the traditional ResNet on ImageNet. It also gave SOTA results for Alzheimer's detection \cite{wang2019}.
        
        We also propose a tiny version of DenseNet121, named \textbf{tiny-DenseNet}. To build this model, we analyzed the internal latent representations of DenseNet121 trained on Dx. We first computed the similarity matrix between each layer of a trained DenseNet using the SVCCA \cite{raghu2017svcca} algorithm and we observed a strong correlation between the blocks 2 and 3 (see Supplementary). We thus removed the \nth{3} block and we decreased the growth rate from $k=32$ to $k=16$ to have a network about 10 times smaller than the original DenseNet121 and with a size comparable with tiny-VGG (892K parameters for tiny-VGG vs 1.8M parameters for tiny-DenseNet).
    
    \subsection{Comparison to Regularized Linear Models}
        
        For comparison purposes, we also evaluated the performance of $\ell_2$-regularized regression and logistic regression (for classification) on the 3 target tasks as it has been shown to perform comparably with kernel methods and CNN models \cite{schulz2020, abrol2020} (at least for phenotype prediction in the small data regime). The penalty term $\alpha \in\{10^{-2}, 10^{-1}, 1\}$ is tuned by grid-search on the validation set.
    
    \subsection{2D slice-level CNN}
        
        For completeness, we compared the 2D-slice level approach against its 3D counterpart. Specifically, we performed the same experiments in the small data regime ($N=500$) with 2D ResNet, DenseNet and VGG. Each 3D scan is decomposed into chunks of 3 consecutive axial slices. The detailed experiments and results can be found in the Supplementary and they are discussed section \ref{discussion_sec}.
    
    \subsection{Metrics, loss functions and optimizer}
        For binary classification tasks, we always reported the Area Under Curve-ROC (AUC) as a reference metric to compare the models. It does not depend on the threshold of the classifier and it allows to compare only the discriminating power of the networks. The balanced accuracy (mean between sensitivity and specificity) has also been reported. For age prediction, we reported both the $\ell_1$ (Mean Absolute Error or MAE) and $\ell_2$ (Mean Squared Error) errors as well as the Pearson correlation coefficient $r$ between the true age $y$ and the predicted age $\hat{y}$ and the coefficient of determination $R^2$ obtained with a linear regression of $y$ vs $\hat{y}$. Since we wanted to be robust to outliers, we used the $\ell_1$ loss for age prediction. As for sex prediction and Dx, we simply used the Binary Cross-Entropy (BCE) loss and, since the dataset was balanced for these 2 classification problems, we did not weight our loss. 
    
    \subsection{Cross-Validation Strategy}
        \label{cross_val_strategy}
        In order to report the scores on the independent test set BSNIP with different training sample sizes, we performed Repeated Learning-Testing (RLT) \cite{arlot:hal-00407906}, similarly to \cite{abrol2020, schulz2020}. 
        RLT is sometimes also referred to as Monte-Carlo Cross-Validation (MTCV), even if in  MTCV one could theoretically use the same training split more than once \cite{arlot:hal-00407906}.
        Specifically, for a given training set size $N\in \{100, 300, 500, 10^3, 1600, 10^4\}$, we randomly picked $N$ training samples among $N_{tot}$ ($N_{tot}=10420$ for age/sex prediction and $N_{tot}=605$ for Dx), stratified on the label to predict (in order to avoid any bias during the sampling). The left-out set containing $N_{tot}-N$ samples is used for validation for Dx. The independent set BIOBD \cite{hozer2020biobd} is used for validation for age/sex prediction. We repeated this procedure 10$\times$ for $N<500$, 5$\times$ for $500 \le N < 10^4$  and 3$\times$ for $N=10^4$. We used BSNIP as independent test set and we reported in Fig.~\ref{fig:learning_curves_bsnip} the averaged results with the corresponding standard deviation. For each repetition, we constantly sampled the same training/validation sets for the different models. Please note that we chose RTL instead of k-fold stratified CV since we wanted to fix the number of training samples at each run.
    
    \subsection{Learning Curves and Convergence Speed}
        \label{meth_learning_curves_convrgence}
        For the 3 tasks at hand, we compared the performances of the CNN architectures, described above, as the training size varied from $N=100$ to $N=10^4$ (resp. $N=500$) for age and sex prediction (resp. Dx). Importantly, we tested the models on the independent test set BSNIP with a constant size of $N=200$ (resp. $N=394$). In order to have a fair comparison, we did not perform any data augmentation in these experiments.
        
        We used an early-stopping criteria to stop the training based on the validation loss. Specifically, let $(\hat{\sigma}_k^{(n)})^2 = \frac{1}{k}\sum_{i=n+1}^{n+k} (\mathcal{L}_i - \bar{\mathcal{L}}_{n,k})^2$ be the rolling window variance of the validation loss $\mathcal{L}$ over $k$ epochs where $\bar{\mathcal{L}}_{n,k}=\frac{1}{k}\sum_{n=n+1}^{n+k}\mathcal{L}_n$. We stop the training at epoch $n$ 
        when $\hat{\sigma}_k^{(n)} < \epsilon$. We fixed $k=20$ and $\epsilon=0.6$ for age prediction (i.e., 7.2 months), and $\epsilon=0.05$ for sex prediction and Dx. Intuitively, this means that we consider that the network has converged if the validation loss remains stable for the next $k$ iterations (without improvement or deterioration). In practice, setting bigger $k$ did not change the stopping epoch $n$ (see figures \ref{fig:age_N_500_cnn_losses} and \ref{fig:dx_N_500_cnn_losses} in the Supplementary).
        
        In addition, we also studied the convergence speed of our networks for each task according to the number of training samples $N$. To do this, we reported the number of iteration steps until convergence (defined above) for each network, task and $N$ for both quasi-raw and VBM pre-processing (see Table~\ref{tab:overall_results_sex_age_dx_archi_N500} and figure \ref{fig:convergence_speed} in the Supplementary).

    \subsection{Data Augmentation Strategies}
        \label{data_augmentation_strats}
        
        \begin{figure*}[h]
            \centering
            \includegraphics[width=\linewidth]{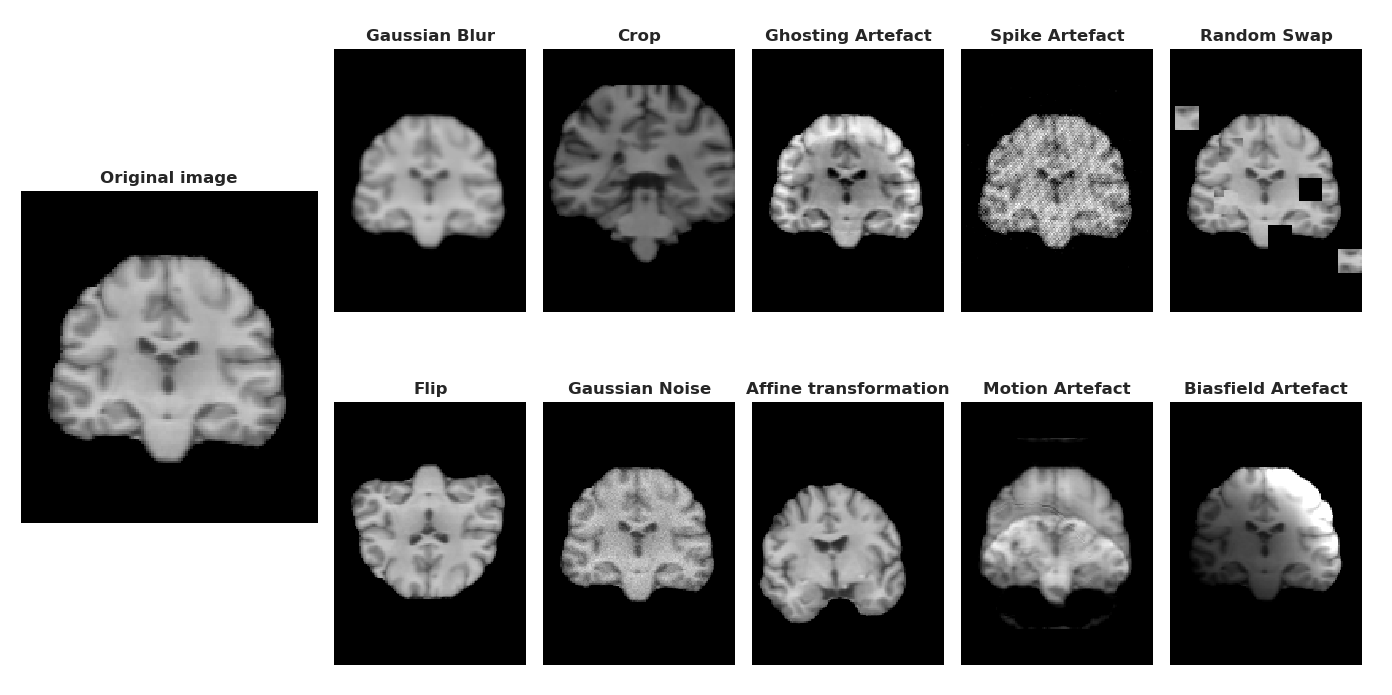}
            \caption{Illustration of the data augmentation techniques applied to quasi-raw data. The same transformations have been applied to VBM data on-the-fly during training (except for MRI artifacts)}
            \label{fig:data_augmentations_quasi-raw_example}
        \end{figure*}
        
        As pointed out in table \ref{comparison_neuroimage_cnn}, we tested various data augmentation strategies. We considered 5 classical transformations widely used in computer vision: \textbf{random flips}, \textbf{Gaussian blur}, \textbf{Gaussian noise}, \textbf{random crop} and \textbf{affine transformations}. All of them are supposed to preserve the semantic information inside the images, while imposing stronger geometric invariance to the final trained CNN model.
        Furthermore, we also evaluated recent augmentation techniques \cite{shaw2019_motion_artefact, zhuo2006_MRI_artefacts, van1999} specifically designed for MRI data: \textbf{ghosting artefacts} \cite{zhuo2006_MRI_artefacts}, \textbf{spike artefact} \cite{zhuo2006_MRI_artefacts}, \textbf{bias-field artefact} \cite{van1999} and \textbf{motion artefact} \cite{shaw2019_motion_artefact}. 
        Finally, we also tested \textbf{swapping} \cite{chen2019self_supervision} which has been originally introduced in the context of self-supervision for both classification and segmentation tasks, in particular on MR images. It consists in picking 2 patches at random location in the image and swapping them. This procedure is repeated $20$ times. Originally, a decoder was added to the network so that it could restore back the initial image $x$ based on the context surrounding each misplaced patch. Here, as we did not want to modify the architecture, we directly perform the downstreaming task (age prediction, sex prediction or Dx) by giving the transformed image $\tilde{x}$ to the CNN. The network should implicitly learn the anatomical features of $x$ given $\tilde{x}$ to perform the downstreaming task.
        
        All of these transformations, along with their hyper-parameters, are detailed in table \ref{tab:data_augmentation_description} in the Supplementary. They have all been applied on-the-fly during training with a probability of $p=0.5$ for each input scan. The test set was never transformed and we did not apply Test-Time Augmentation \cite{VGG_Simonyan} as the network should be already invariant to the transformations applied during training. We propose to assess the importance of each data augmentation technique separately using either VBM or quasi-raw data for the 3 tasks. To the best of our knowledge, this is the first time MRI artefacts are employed for data augmentation for age prediction, sex classification and Dx. 
        Please note that we applied MRI artefacts only on quasi-raw images and not on VBM data since they were conceived for images and not for gray matter density maps. Indeed, in order to apply MRI artefacts, one needs to compute the inverse Fourier transform to map the image back to the k-space \cite{zhuo2006_MRI_artefacts}. When considering VBM data, one would also need to compute the backward mapping from gray matter density to the original image and this would be computationally too demanding and prone to error.

    \subsection{Deep Ensemble Learning}
        \label{deep_ensemble}
        
        Uncertainty quantification in DL is very important, especially for clinical applications. In \cite{lakshminarayanan2017}, authors introduced deep ensemble learning as a simple method to integrate both aleatoric uncertainty (related to the noise in the data) and epistemic uncertainty \cite{gal2016phd, kendall2017uncertainties} (associated to the model's uncertainty). It consists in training independently $T$ identical DNN with different starting points $(w_t^0)_{t\in [1,..T]}$ and shuffling the data during the stochastic gradient descent optimization step. At the end of the optimization, this gives $T$ models $f_{\hat{w_t}}$ where each model's weights $\hat{w_t}$ can be seen as a sample of an approximation of the highly multi-modal distribution $p(w|X,Y)$ where $(X,Y)$ represents the training set (more details in \cite{gustafsson2020bench}). Usually, for classification, $f_w(x)$ is the output after the softmax layer, giving a probability vector. For regression, $f_w(x)$ can be modelled as a Gaussian distribution whose mean and variance (representing the aleatoric heteroscedastic uncertainty \cite{kendall2017uncertainties}) are learnt during training by optimizing the log-likelihood \cite{lakshminarayanan2017, gustafsson2020bench}. Here, as we want to study the small data regime and we are interested in the epistemic uncertainty, we fix the variance for regression and we do not optimize it\footnote{We also observed a strong over-fitting effect when we tune the variance on age prediction with $N=500$.}. Thus, $f_w(x)$ outputs only one value. With the proposed framework, the final prediction for an input image $x$ is given by:
        \begin{itemize}
            \item for classification: $\hat{p}(y|x) = \frac{1}{T}\sum_{t=1}^T f_{\hat{w_t}}(x)[y]$
            \item for regression: $\hat{p}(y|x) = \mathcal{N}(y; \hat{\mu}(x), \hat{\sigma}^2(x))$  with $\hat{\mu}(x)=\frac{1}{T}\sum_{t=1}^T f_{\hat{w}_t}(x)$ and $\hat{\sigma}^2(x) = \frac{1}{T}\sum_{t=1}^T (f_{\hat{w}_t}(x)-\hat{\mu}(x))^2$
        \end{itemize}
    
        As pointed out in \cite{gustafsson2020bench}, Monte-Carlo Dropout \cite{gal2016dropout, gal2016phd} could be another simple way to quantify aleatoric and epistemic uncertainties. It has been successfully applied in the medical imaging field to diabetic retinopathy diagnosis \cite{filos2019systematic,leibig2017leveraging}. However, it has been shown to under-perform compared to Deep Ensemble Learning on various real-world computer vision tasks \cite{gustafsson2020bench}. As a result, we employed the latter method. Please note that we also evaluated Monte-Carlo Dropout by integrating Concrete Dropout \cite{gal2017concrete_dropout} in our models (see Supplementary). The results are discussed in section \ref{discussion_sec}.
        
        In practice, we evaluated the calibration error of our CNN models within the Deep Ensemble learning framework to quantify their predictive uncertainty. Briefly, a well calibrated classifier should give a probability for a given class equals to its occurrence's probability.  A mis-calibrated model indicates that it makes under or over-confident predictions. It is usually measured by the Expected Calibration Error (ECE) that gives the confidence error between a perfectly calibrated model and the model at hand. This metric can be extended to regression with the Area Under Calibration Error (AUCE) score as introduced in \cite{gustafsson2020bench} (see Supplementary for more details).

       \subsection{Optimization and Implementation Details} 
       
       Input data from both pre-processing were normalized to have zero mean and unit variance. Furthermore, we use the optimizer Adam \cite{kingma2014adam} with the default parameters $\beta_1=0.9$ and $\beta_2=0.999$ and we set the learning rate to $\alpha=10^{-4}$ decreasing it by a constant factor $\gamma$ every 10 epochs. This factor is tuned between $\{0.2, 0.5, 0.9\}$ through cross-validation for each task and training set size. We also set a batch size $b=8$ to limit the computational cost for $N < 10^4$ and $b=32$ for $N=10^4$. 
        All CNN architectures are implemented with Pytorch v1.6 \cite{paszke2019pytorch} while the linear models with scikit-learn v0.23 \cite{pedregosa2011scikit}. Finally, the experiments with $N=500$ samples were all performed on a single Quadro RTX 8000 GPU with 48GB. The code is available \href{https://github.com/Duplums/bhb10k-dl-benchmark}{here}.

\section{Results}

    \subsection{CNN Performance on Dx, Age and Sex Prediction at N=500}
        \label{cnn_perf_N500}
        \begin{table*}[h!]
            \centering
            \resizebox{\textwidth}{!}{\begin{tabular}{|c|c|c||c|c|c|c||c|c||c|c||c|c|}
                \rowcolor{Gray}
                \multicolumn{1}{c|}{\textbf{Preprocessing}} & \multicolumn{2}{c||}{\textbf{Architecture}} &  \multicolumn{4}{c||}{\textbf{Age}} & \multicolumn{2}{c||}{\textbf{Sex}} & \multicolumn{2}{c||}{\textbf{Dx}} & \multicolumn{2}{c}{\textbf{Resources}}\\
                \hline
                \cline{3-13} 
                & Model & \#Params & MAE $\downarrow$ & RMSE $\downarrow$ & $r$ $\uparrow$ & $R^2$ $\uparrow$ & AUC $\uparrow$ & BAcc $\uparrow$ & AUC $\uparrow$ & BAcc $\uparrow$ & Time & GPU\\
                \cline{2-13}

                \multirow{8}{*}{VBM} & \textit{Linear Model} \footnote{The input images contain $2.1$M voxels however since a brain mask is applied, only 400K voxels are used to train the model here. }& 400K & $7.19 \pm 0.17$ & $8.56\pm 0.17$ & $0.78 \pm 0.014$ & $0.61\pm 0.02$ & $\mathbf{0.93 \pm 0.01} $ & $0.82\pm 0.02$ & $0.78\pm 0.007$ & $0.71\pm 0.01$ & - & -\\ 
                                        & DenseNet121 \cite{DenseNet_Huang} & 11.2M & $\mathbf{6.02 \pm 0.24}$ & $\mathbf{7.25 \pm 0.34}$ & $\mathbf{0.83 \pm 0.008}$ & $\mathbf{0.70 \pm 0.01}$ & $0.91 \pm 0.01$ & $\mathbf{0.83 \pm 0.01}$ & $0.78 \pm 0.01$ & $0.72 \pm 0.02$ & 11s & 10GB \\
                                                    & ResNet18 \cite{ResNet_He} & 33.1M & $6.91 \pm 0.52$ & $8.47 \pm 0.64$ & $0.81 \pm 0.02$ & $0.66 \pm 0.03$ & $0.91\pm 0.01$ & $0.83 \pm 0.01$ & $0.78 \pm 0.01$ & $0.71 \pm 0.01$ & 5s & 4.6GB \\ 
                                                    & ResNet34 \cite{ResNet_He} & 63.4M  & $7.14 \pm 0.28$ & $8.87 \pm 0.38$ & $0.75 \pm 0.03$ & $0.57 \pm 0.05$ & $0.91 \pm 0.01$ & $0.80 \pm 0.02$ & $ 0.75 \pm 0.005$ & $0.69 \pm 0.01$ & 8s & 5.4GB \\
                                                    & ResNet50 \cite{ResNet_He} & 46.1M & $6.26 \pm 0.29$ & $7.42 \pm 0.3$ & $0.83 \pm 0.02$ & $0.70 \pm 0.03$ & $0.91 \pm 0.008$ & $0.81 \pm 0.02$ & $0.79 \pm 0.009$ & $0.71 \pm 0.01$ & 7s & 9GB \\ 
                                                    & VGG11 \cite{VGG_Simonyan} & 50.1M & $7.03 \pm 0.44$ & $8.63 \pm 0.56$ & $0.78 \pm 0.02$ & $0.62 \pm 0.03$ & $0.80 \pm 0.02$ & $0.72 \pm 0.015$ & $0.72 \pm 0.009$ & $0.66 \pm 0.03$ & 23s & 26GB \\
                                                    & tiny-VGG \cite{cole2017predicting} & 892K & $6.94 \pm 0.12$ & $8.21 \pm 0.20$ & $0.78 \pm 0.02$ & $0.61 \pm 0.04$ & $0.91 \pm 0.005$ & $0.82 \pm 0.007$ & $0.79 \pm 0.008$ & $0.71 \pm 0.01$ & 7s & 4.7GB \\
                                                    & tiny-DenseNet & 1.8M & $6.43 \pm 0.22$ & $7.74 \pm 0.24$ & $0.81 \pm 0.02$ & $0.66 \pm 0.03$ & $0.88 \pm 0.01$ & $0.81 \pm 0.02$ & $\mathbf{0.79 \pm 0.01}$ & $\mathbf{0.72 \pm 0.01}$ & 6s & 7GB \\
                                                    & ResNeXt \cite{ResNeXt_Xie} & 25.8M & $6.24 \pm 0.21$ & $7.47 \pm 0.16$ & $0.83 \pm 0.007$ & $0.68 \pm 0.01$ & $0.88 \pm 0.007$ & $0.79 \pm 0.02$ & $0.77 \pm 0.01$ & $0.70 \pm 0.009$ & 28s & 10GB \\
                                                    & SFCN \cite{peng2021} & 2.9M & $6.60 \pm 0.35$ & $8.07 \pm 0.37$ & $0.77 \pm 0.03$ & $0.60 \pm 0.03$ & $0.85 \pm 0.02$ & $0.76 \pm 0.02$ & $0.77 \pm 0.015$ & $0.69 \pm 0.01$ & 6s & 10.6GB \\
                \hline
                \hline
                \multirow{8}{*}{Quasi-Raw}      & \textit{Linear Model} & 400K & - & - & - & - & $0.60\pm 0.03$ & $0.50\pm 0.001$ & $0.51\pm 0.02$ & $0.50\pm 0.002$ & - & -  \\  
                                                & DenseNet121 \cite{DenseNet_Huang} & 11.2M & $10.73 \pm 2.65$ & $13.36\pm 3.01$ & $0.53 \pm 0.05$ & $0.28 \pm0.05$ & $0.81 \pm 0.01$ & $\mathbf{0.68 \pm 0.05}$ & $\mathbf{0.72 \pm 0.01}$ & $\mathbf{0.65 \pm 0.02}$ & 11s & 10GB \\
                                                    & ResNet18 \cite{ResNet_He} & 33.1M  & $7.91 \pm 0.42$ & $9.82 \pm 0.55$ & $0.69\pm 0.03$ & $0.48 \pm 0.04$ & $\mathbf{0.84 \pm 0.01}$ & $0.63\pm 0.02$ & $0.66\pm 0.01$ & $0.63 \pm 0.02$ & 5s & 4.6GB \\ 
                                                    & ResNet34 \cite{ResNet_He} & 63.4M & $8.33 \pm 1.05$ & $10.31 \pm 1.25$ & $0.71 \pm 0.03$ & $0.51 \pm 0.04$ & $0.84 \pm 0.02$ & $0.60 \pm 0.02$ & $0.66 \pm 0.03$ & $0.61 \pm 0.02$ & 8s & 5.4GB \\
                                                    & ResNet50 \cite{ResNet_He} & 46.1M & $12.9 \pm 2.53$ & $15.50 \pm 2.73$ & $0.59 \pm 0.04$ & $0.34 \pm 0.05$ & $0.80 \pm 0.01$ & $0.62 \pm 0.04$ & $0.66 \pm 0.009$ & $0.61 \pm 0.01$ & 7s & 9GB \\ 
                                                    & VGG11 \cite{VGG_Simonyan} & 50.1M & $16.42\pm 3.23$ & $18.83 \pm 3.09$ & $0.69 \pm 0.02$ & $0.48 \pm 0.03$ & $0.68 \pm 0.05$ & $0.54 \pm 0.02$ & $0.61\pm 0.02$ & $0.53 \pm 0.04$ & 23s & 26GB \\
                                                    & tiny-VGG \cite{cole2017predicting} & 892K & $9.84 \pm 0.48$ & $12.08 \pm 0.66$ & $0.66\pm 0.01$ & $0.43 \pm 0.02$ & $0.76\pm 0.03$ & $0.65 \pm 0.05$ & $0.69 \pm 0.01$ & $0.62 \pm 0.01$ &  7s & 4.7GB\\
                                                    & tiny-DenseNet & 1.8M & $14.7 \pm 4.19$ & $20 \pm 8.0$ & $0.52\pm 0.07$ & $0.27 \pm 0.07$ & $0.76 \pm 0.03$ & $0.62 \pm 0.06$ & $0.68 \pm 0.02$ & $0.62 \pm 0.01$ & 6s & 7GB \\
                                                    & ResNeXt \cite{ResNeXt_Xie} & 25.8M & $13.30 \pm 2.2$ & $16 \pm 2.32$ & $0.56\pm 0.05$ & $0.32 \pm 0.05$ & $0.78\pm0.02$ & $0.61\pm 0.02$ & $0.64 \pm 0.01$ & $0.60 \pm 0.007$ & 28s & 10GB \\
                                                    & SFCN \cite{peng2021} & 2.9M & $\mathbf{6.68 \pm 0.23}$ & $\mathbf{8.40 \pm 0.30}$ & $\mathbf{0.75 \pm 0.02}$ & $\mathbf{0.56 \pm 0.05}$ & $0.75 \pm 0.01$ & $0.53 \pm 0.02$ & $0.70 \pm 0.008$ & $0.64 \pm 0.02$ & 6s & 10.6GB\\
                \hline
                \end{tabular}}
                \caption{Comparison between several architectures and preprocessing at N=500 samples in the training set without data augmentation. The results are reported on BSNIP, an independent test set, with a 5-fold RLT. MAE=Mean Absolute Error, RMSE=Root Mean Squared Error, $r$=Pearson correlation coefficient, $R^2$=coefficient of determination, AUC=ROC-Area Under Curve, BAcc=Balanced Accuracy, Time=GPU time over 1 epoch for age prediction and GPU=GPU memory usage during training for age prediction.}                \label{tab:overall_results_sex_age_dx_archi_N500}
        \end{table*}
        
        Table \ref{tab:overall_results_sex_age_dx_archi_N500} summarizes the results with the architectures presented in section \ref{cnn_archi_details} and using $N=500$ training samples within a 5-fold RLT strategy described in section \ref{cross_val_strategy}. We also reported the performance of 2D CNN models in the Supplementary.  
        
        First, we acknowledge that 3D CNN models performed always comparably or better than their 2D counterpart. We chose to always keep this 3D approach. Second, while all the networks performed very similarly on all tasks for the VBM pre-processing, a strong over-fitting effect has been observed on quasi-raw data, independently from the depth or size of the networks. Notably, SFCN outperforms the other CNN on age prediction with quasi-raw data thanks to its dropout layer. However, it under-performs on the other two tasks. Based on these results, we provide an in-depth study on 4 of these networks, namely tiny-VGG, tiny-DenseNet, DenseNet121 and ResNet34, as representative of the main CNN families. We did not retain ResNeXt (Inception-ResNet family) for computational reasons (it takes $3\times$ much time than DenseNet for a feed-forward pass) and because it gave very similar results with the other networks. 
    
    
    \subsection{Learning Curves}
    \label{learning_curves_results}
    
    \begin{figure*}
        \centering
        \includegraphics[width=\linewidth]{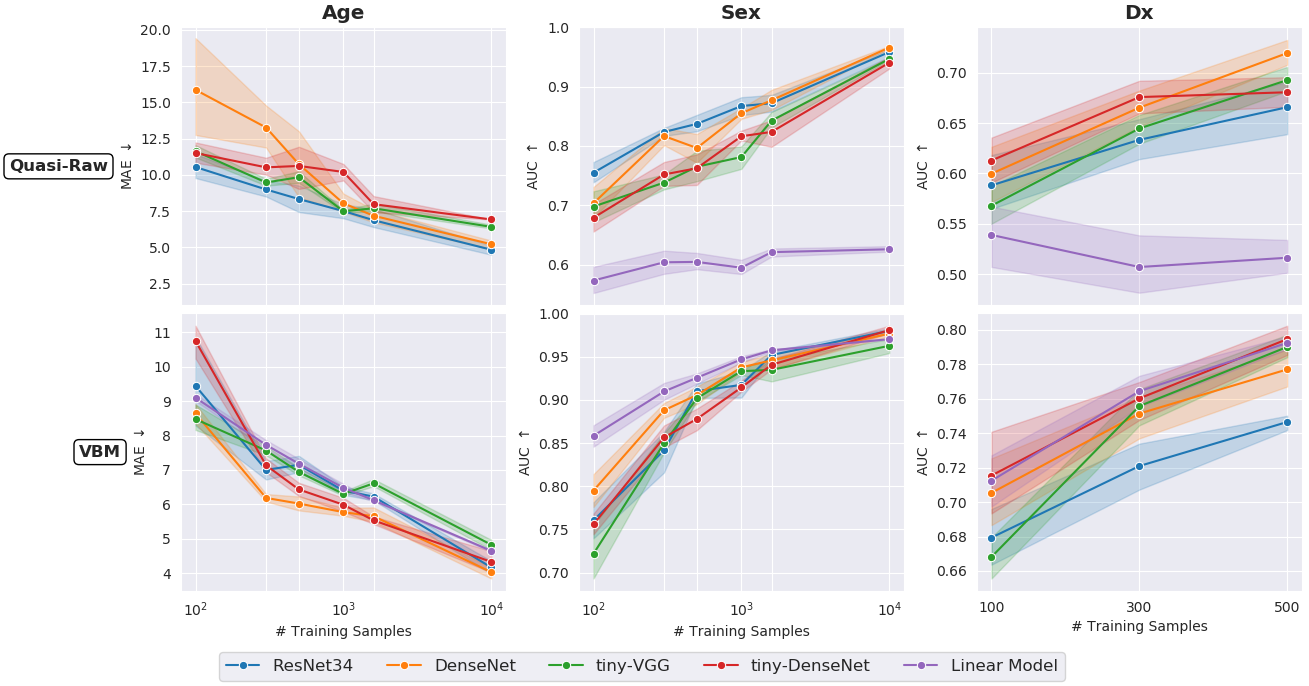}
        \caption{Learning curves of CNN models with quasi-raw and VBM pre-processings. Overall, CNNs outperform linear model with quasi-raw data but they perform similarly with their linear counter-part only with VBM images. DenseNet offers a good compromise since it performs equally well in the small data regime on Dx and in the very big data regime $N=10^4$ on age and sex prediction, for both VBM and quasi-raw images. The results are reported on the independent data-set BSNIP (images were not acquired with the same acquisition protocol than images in train). The results on age prediction with a linear model and quasi-raw images are outside the plot and thus not reported ($MAE>20$ for all $N\in [100, 10^4]$).}
        \label{fig:learning_curves_bsnip}
    \end{figure*}
    
    We have reported the performances of the neural networks as well as of the 
    linear model in figure \ref{fig:learning_curves_bsnip}. For age and sex prediction, DenseNet and ResNet always perform better than the linear model, no matter the pre-processing, but given enough data ($N\gg 10^3$). Bigger models (\textit{e.g} DenseNet) also perform better on quasi-raw data when $N=10^4$ than their smaller counterpart (\textit{e.g} tiny-DenseNet). Importantly, as before, we generally observed a drop in performance when using quasi-raw data as opposed to VBM. We will investigate more this difference in generalization in the next section. 
    
    For completeness, we also plotted the convergence plots of the CNN architectures in the supplementary (Fig. \ref{fig:convergence_speed}). We used the same convergence criterion for all networks and tasks, as defined in \ref{meth_learning_curves_convrgence}. 
    
    \subsection{Site Effect}
    
    \begin{figure*}[!ht]
        \centering
        \includegraphics[width=.9\linewidth]{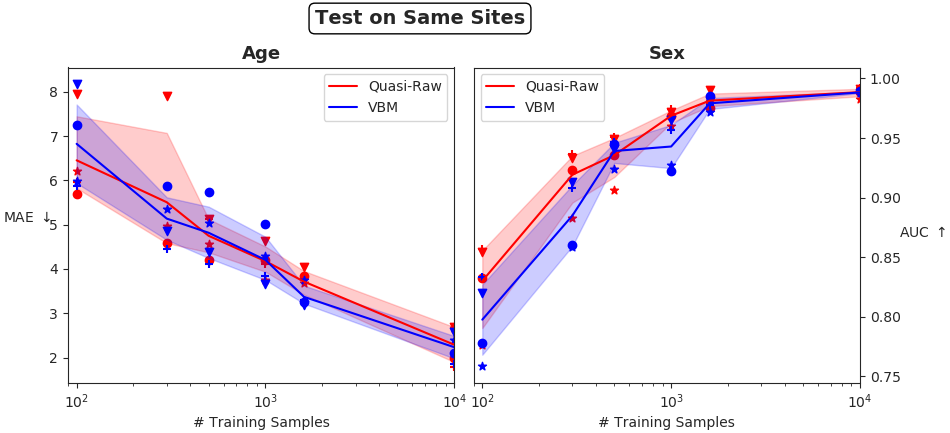}
        \includegraphics[width=.9\linewidth]{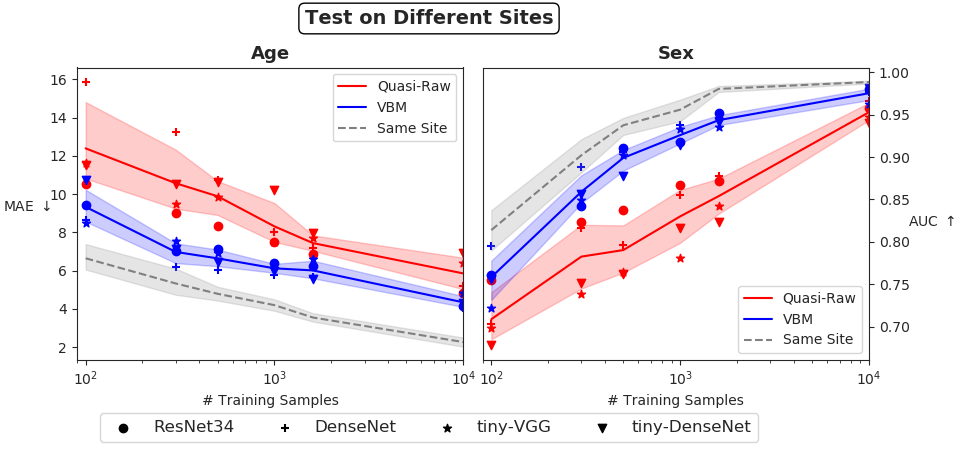}
        \caption{Performance of CNN models when trained with a varying number of training data on BHB-10K and tested i) on images coming from the same acquisition sites (top); or ii) on images coming from different sites (bottom). VBM pre-processing reduces the bias induced by acquisition site, especially in the small data regime ($N<10^3$). There is no difference in performance between quasi-raw and VBM data when testing on images coming from the same sites as the training images (\textit{in-site}). The gray dashed lines in the bottom images represent the average performance on \textit{in-site} images (across pre-processing and CNN) and it was reported to ease comparison. }
        \label{fig:site_effect}
    \end{figure*}
    
    In figure \ref{fig:site_effect}, we plotted the performances of the CNN models when they are evaluated on images coming from the same acquisition sites they are trained on (\textit{in-site}) or different ones (\textit{out-site}). Two main effects can be observed. First, there is a strong gap between performance on \textit{in-site} and \textit{out-site} images, even when CNNs are trained on a big multi-site data-set ($+3.5$ MAE $p<10^{-3}$ for age prediction, $-3.6\%$ AUC $p<10^{-3}$ for sex prediction at $N=10^4$ training samples for quasi-raw images). Second, for \textit{out-site} images, CNN models perform significantly better with VBM data than quasi-raw data when $N \le 10^3$, in line with results table \ref{cnn_perf_N500} ($+2.2$ MAE $p<0.01$, $-9\%$ AUC $p<0.003$). However, it is not the case for \textit{in-site} images where CNNs perform similarly.

    \subsection{Data Augmentation}

        \begin{figure*}[h]
            \centering
            \includegraphics[width=\linewidth]{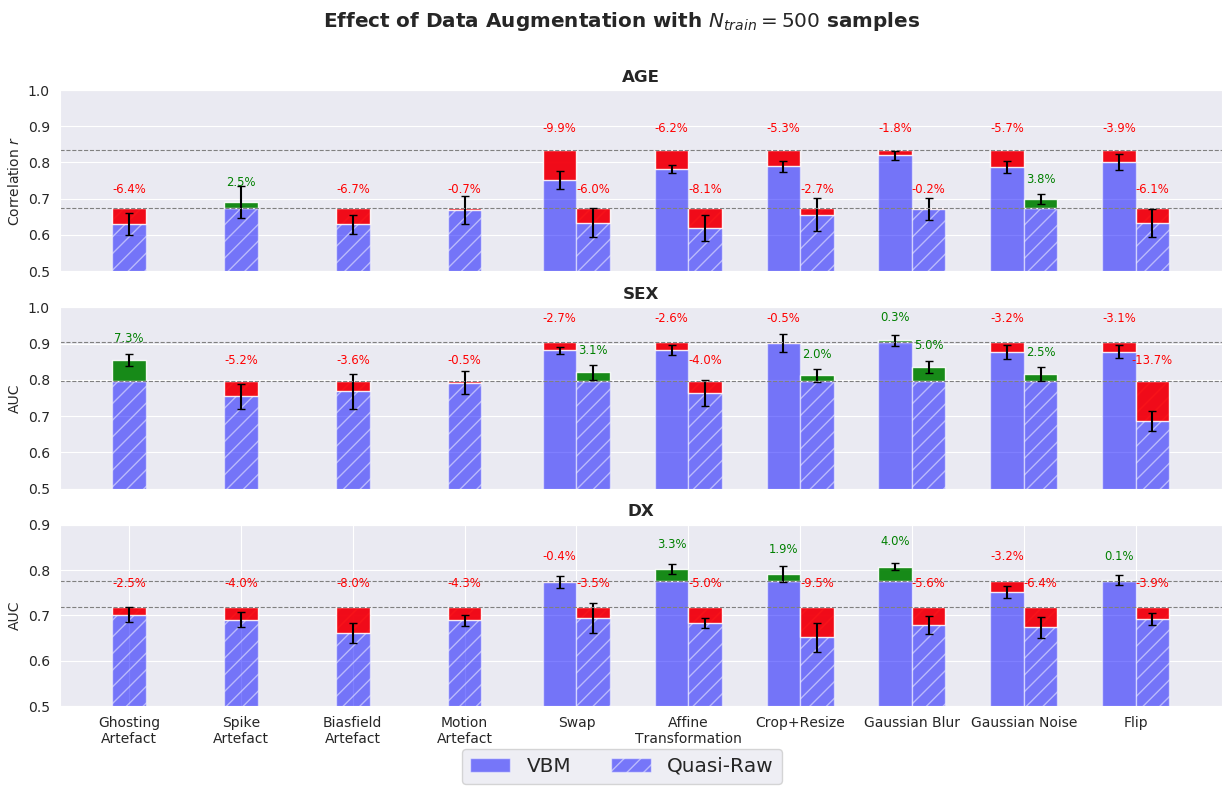}
            \caption{Current data augmentation (D.A) techniques are highly task- and pre-processing-dependent. It does not result in large improvement and, overall, it even degrades the performance for both VBM and quasi-raw images. The error bars are obtained using a 5-split RLT strategy using each time only one data augmentation strategy. We reported the results obtained on the independent test set BSNIP ($N_{HC}=200, N_{SCZ}=194$). The black dashed lines are the baselines without D.A.}
            \label{fig:data_augmentation}
        \end{figure*}
        
        In figure \ref{fig:data_augmentation}, we showed the results when using data augmentation (D.A) for the 3 tasks with $N_{train}=500$ samples (small real-life data regime). We only tested the usefulness of each strategy alone and not their combination, since this would have been computationally too demanding. Here, we only used DenseNet since it performs well on all tasks, except for age prediction with quasi-raw data (see fig. \ref{learning_curves_results}). In that case, we trained ResNet34 because it was much more stable. Overall, D.A. seems to be counter-productive for age prediction. MRI artefacts have no benefit for diagnosis prediction and they only help for sex prediction when using quasi-raw data. Classical affine transformations, crop+resize and Gaussian blurring have a positive impact only when using VBM data for Dx. 
    
    \subsection{Deep Ensemble Learning}
    
    \begin{figure*}[h!]
        \centering
        \includegraphics[width=\linewidth]{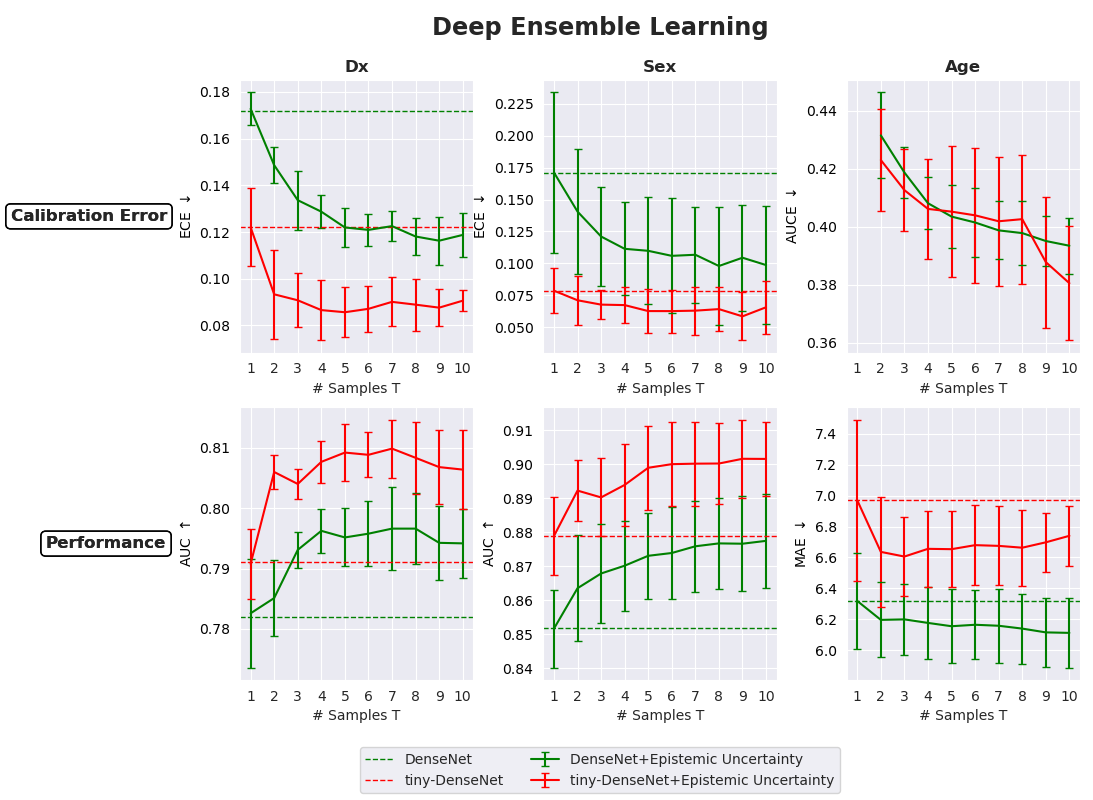}
        \caption{Performance of big and small networks (resp. DenseNet and tiny-DenseNet) as we increase the number of ensemble models $T$ when $N=500$ training samples. tiny-DenseNet is always better calibrated than DenseNet while also performing better in the small data regime. AUC=Area Under ROC Curve, MAE=Mean Absolute Error, ECE=Expected Calibration Error, AUCE=Area Under Calibration Error.}
        \label{fig:deep_ensemble_performance}
    \end{figure*}

     In figure \ref{fig:deep_ensemble_performance}, we reported the performances and calibration errors of DenseNet and tiny-DenseNet on the 3 tasks as we increase the number of independently trained models $T$ (see section \ref{deep_ensemble}) with $N_{train}=500$ training samples (small data regime). We have only used the VBM data to perform this analysis, considering the stability of CNN with this pre-processing. The baselines (dashed lines) are given with the deterministic version of DenseNet and tiny-DenseNet (\textit{i.e,} $T=1$). 
            
 \section{Discussion}
    \label{discussion_sec}
    \subsection{Extensive CNN Comparison with $N=500$}
    Overall, we did not found significant differences between CNNs in the small data regime with VBM images, no matter the depth or the architecture, for the 3 tasks. This is somewhat expected since very deep architectures have been introduced originally for very large-scale datasets (\textit{e.g} ImageNet with millions of 2D images). Also, even wider models such as ResNeXt (based on the idea of ResNet with Inception modules) did not help to improve the performance neither on VBM data nor on Quasi-Raw data. However, we noticed a strong over-fitting effect when using quasi-raw images for the 3 tasks and for every network, resulting in a strong drop in performance (-7\% AUC on sex prediction between the 2 best models, -6\% AUC for Dx and -8\% correlation for age prediction). Notably, SFCN performed well on age prediction with quasi-raw data, matching the performance on VBM images (in line with \cite{peng2021}). This can be attribute to its last dropout layer. However, it still under-performed for the other 2 tasks. As we shall demonstrate, this over-fitting effect can be largely attributed to the bias towards the acquisition sites (in line with \cite{Jonsson2019}), which explains why it has not been systematically reported \cite{cole2017predicting, wen2020}.
    
    Interestingly, we noticed that 2D models perform significantly worse than their 3D counterpart for all tasks, in line with \cite{wen2020} for Alzheimer's detection (see table \ref{tab:2d_cnn_perf_N500} in the Supplementary). It means that 3D CNN models, while being computationally more expensive during training, benefit from the underlying 3D anatomical structure of the brain that a 2D approach cannot capture, even in the small data regime. 
    
    Ultimately, the resources taken by all 3D networks, VGG11 and ResNeXt excepted, were comparable in terms of GPU time. VGG11 was very time-consuming because of its 3 Fully-Connected layers at the end (replaced by a single Fully-Connected layer, lighter, in tiny-VGG). ResNeXt is very large, even if it has less parameters than its ResNet50 counterpart, and it remains computationally very costly.

    \subsection{Scaling up to $N=10^4$}
    
     The learning curves reported in figure \ref{learning_curves_results} confirmed the over-fitting effect observed with $N=500$ training samples and quasi-raw data. With the extensive VBM pre-processing, the networks performed better than quasi-raw no matter the data regime (from $N=100$ up to $N=10^4$, while being greatly reduced for $N \gg 10^3$, see figure \ref{fig:site_effect}) or the task. Nonetheless, we also noticed that big networks (DenseNet121, ResNet34 with more than 10M parameters) gave better results when $N=10^4$ with quasi-raw data (-1.7 MAE, $p<10^{-3}$ for age prediction, +2\% AUC, $p<10^{-2}$ on sex prediction between tiny-DenseNet and DenseNet). As a result, adding more parameters (from 11M for DenseNet up to 60M for ResNet) is useful only when dealing with quasi-raw data in the very large data regime. 
     
     As reported previously \cite{abrol2020}, big CNN models outperform linear models in the large-scale data regime when $N=10^4$ with VBM data (-0.7 MAE for age prediction $p<0.003$, +0.7\% AUC for sex prediction $p<10^{-2}$). It was also expected that linear model gives random results (i.e., balanced accuracy around 0.5) when using quasi-raw data without feature extraction. However, we should remark that it performed very well in the small data regime when $N \le 10^3$ on all tasks, given an appropriate pre-processing (VBM here), in line with \cite{wen2020}. For instance, it performed similarly to all SOTA CNN architectures for Dx up to 500 samples (79\% AUC with $N=500$, matching tiny-DenseNet or tiny-VGG). Overall, big networks are required when dealing with a big minimally pre-processed MRI data-set to reach SOTA results.

     \subsection{Site Effect}
     
    We should emphasize that we did not retrieve the very accurate results obtained on UKBioBank with $N\ge 10^4$ in \cite{peng2021} for age and sex prediction (resp. $MAE=2.1$ and $Acc=99\%$ with quasi-raw data). As shown in figure \ref{fig:site_effect}, this is largely due to the underlying site effect that highly deteriorates the performance of CNN ($+3.5$ MAE $p<10^{-3}$ for age prediction, $-3.6\%$ AUC $p<10^{-3}$ for sex prediction with quasi-raw data and $N=10^4$ between \textit{in-site} and \textit{out-site} images). Indeed, we retrieved the same performances as in \cite{peng2021} when testing the models on \textit{in-site} images. However, there is a drop in performance when using the same models on a \textit{out-site} independent test set. Our results also explain why authors in \cite{wen2020} found no difference between minimal and quasi-raw pre-processing on Alzheimer's detection (the results were reported only on an \textit{in-site} validation set). Even if the extensive non-linear pre-processing indeed removes part of the site-effect, this bias still remains as pointed out figure \ref{fig:site_effect} (difference between gray dashed line and blue line). \\
     
    As opposed to UKBioBank, our data-sets are highly multi-centric and this appears to be critical when performing ML with neuroimaging data (as confirmed by \cite{wachinger2021detect}, with traditional ML and hand-crafted features for age prediction, or by \cite{glocker2019machine} on sex prediction). As highlighted in \cite{Jonsson2019}, exploiting images acquired with the same protocol as the test images is one way to deal with this issue but it assumes to have access to these images beforehand and to fine-tune the model on them (which is impractical in the real clinical setting). Other solutions are also starting to emerge towards debiased DL algorithms by directly accounting for the bias during the optimization \cite{tartaglione2021end}. Even if it is well-known in computer vision that ML and DL algorithms perform poorly when images from train and test sets come from 2 different domains (\textit{i.e} domain gap) \cite{torralba2011unbiased}, this issue is still rarely mentioned when performing a benchmark with neuroimaging data \cite{schulz2020, abrol2021deep}. Here, we demonstrated that even SOTA DL models, trained on a large-scale multi-site data-set, still under-perform on \textit{out-site} images compared to \textit{in-site} images. 
    
     From this perspective, BHB-10K is quite distinctive from UKBioBank for its diversity (images are coming from more than $70$ different sites) and we believe that it can provide a new way to test and benchmark ML and DL algorithms on images from sites not seen during training.

    \subsection{Data Augmentation}
    Overall, we found that data augmentation brings little or no improvement for both VBM and quasi-raw images. As opposed to \cite{peng2021, cole2017predicting, armanious2020age}, affine transformation and flip did not improve the performance on age prediction. Once again, differently from the above-mentioned studies, our results are reported on an independent data-set which may explain the differences.  Interestingly, horizontal and vertical flip degrade significantly the performance only for sex prediction, which is expected since the localization of left and right hemisphere matters to discriminate between male and female. For Dx, we noted some improvements with VBM data when introducing traditional D.A such as affine transformation, crop or Gaussian blur. Please note that we did not smooth the data with a Gaussian kernel during the VBM pre-processing. As a result, this indicates that smoothing is beneficial when dealing with noisy data such as SCHIZCONNECT-VIP (in line with \cite{abrol2020}).
    
    From figure \ref{fig:data_augmentation}, it can be seen that data augmentation is both task- and pre-processing-dependent and it does not necessarily result in large improvement neither for regression nor classification tasks. For an easy task (sex prediction with $AUC \ge 0.9$) it significantly improves the performance only with quasi-raw data (\textit{i.e}, with ghosting artefact or Gaussian blur) while for a hard task such as Dx, it helped only on VBM data. This mitigates the usefulness of current data augmentation techniques on brain MRI, especially when all images have been aligned to the same template and re-sampled to the same spatial resolution. Even with the minimal pre-processing (i.e., quasi-raw), there is no clear improvement with the standard D.A (affine transformation, Gaussian blur, etc.). Furthermore, we also showed that adding MRI artefacts into the data augmentation strategy brings overall no improvement and it actually worsen the results most of the time (except for ghosting artefact and spike artefact for sex and age prediction respectively). 
    
    \subsection{Predictive Uncertainty in the Small Data Regime}
    
    Quantifying the model uncertainty associated to a prediction is very important when dealing with computer-aided diagnosis systems. However, this is rarely mentioned in the literature even if simple calibration metrics exist and have been extended to regression \cite{gustafsson2020bench}. Here, we show in figure \ref{fig:deep_ensemble_performance} that, when using few training samples ($N_{train}=500$), small networks are better calibrated than their bigger counterpart (in line with \cite{guo2017calibration} for vision tasks) and they also perform better. We demonstrated that Deep Ensemble learning provides a simple way to better calibrate the models, no matter their size, while also improving the results (in line with \cite{gustafsson2020bench}). This is not the case for MC-Dropout. Our experiments (see figure \ref{fig:MC_dropout} in the Supplementary) showed that a well calibrated model does not always perform well on the task at hand. For instance, while Bayesian tiny-DenseNet is perfectly calibrated on Dx with $ECE=0.04$ (-8\% compared to tiny-DenseNet), it loses 7\% AUC compared to its deterministic counterpart. This applies to all tasks and models tested, except for age prediction. 
    These results suggest that in usual clinical applications ($N_{train}=500$), it is better to use small networks (\textit{e.g} tiny-DenseNet) with Deep Ensemble learning since it improves both the calibration error and the accuracy.
\section{Conclusion}

Throughout this paper, we have empirically studied several properties of 3D CNN models on neuroimaging data. First, we have shown that all CNN models perform significantly better on VBM data than on quasi-raw images, no matter their architecture. We emphasize this gap is greatly reduced as we scale up to $N=10$k samples. Importantly, we also demonstrated that simple linear models are on par with SOTA CNN on VBM, which suggests that DL model fail to capture non-linearities in the data, as suggested by \cite{schulz2020}. However, this conclusion must be taken with caution since we also shown that CNN models are still very biased towards the acquisition site, even when they are supervised on a highly multi-sites data-set with $N=10$k samples. Extensive non-linear pre-processing such as VBM provides a simple way to limit this bias but it still does not entirely remove it. This effect has also been reported on several other brain MRI data-sets \cite{glocker2019machine, wachinger2021detect} and Chest X-ray data-set with COVID-19 data \cite{tartaglione2021end}. De-biasing methods for DL are starting to emerge mainly in computer vision \cite{tartaglione2021end,kim2019learning, hendricks2018women} but with future potential applications to the neuroimaging field. Suprisingly, we also observed overall no benefits from using data augmentation in the small data regime. In this paper, we showed that it is task- and dataset-specific but a more in-depth study is required and left for future work. Finally, while big CNN models were poorly calibrated on all neuroimaging tasks we trained them on (as also reported on vision tasks \cite{guo2017calibration}), we demonstrated that deep ensemble learning provides a simple and effective way to re-calibrate them, by even improving the performance. We highlight the importance of well-calibrated models in particular for clinical applications. 

As a step towards reproducible research, we made our code publicly available \href{https://github.com/Duplums/bhb10k-dl-benchmark}{here} and we also provide the BHB-10K data-set used throughout this study. We give access to both quasi-raw and VBM data, directly usable within a Python environment for DL (e.g., Pytorch or TensorFlow).

\section*{Data Availability}

Data used throughout this study have been collected through various public platforms (see table \ref{table:demographic_infos} for all the links). We have also engaged a process to release all the pre-processed data-sets publicly available. The releasing status is regularly updated on our Github repository:
\url{https://github.com/Duplums/bhb10k-dl-benchmark}
\section*{Acknowledgements}
This work was performed using HPC resources from GENCI-IDRIS (Grant 2020-AD011011854).

\bibliography{bibliography}

\begin{thebibliography}{10}
\urlstyle{rm}
\expandafter\ifx\csname url\endcsname\relax
  \def\url#1{\texttt{#1}}\fi
\expandafter\ifx\csname urlprefix\endcsname\relax\def\urlprefix{URL }\fi
\expandafter\ifx\csname doiprefix\endcsname\relax\def\doiprefix{DOI: }\fi
\providecommand{\bibinfo}[2]{#2}
\providecommand{\eprint}[2][]{\url{#2}}

\bibitem{alexnet2012}
\bibinfo{author}{Krizhevsky, A.}, \bibinfo{author}{Sutskever, I.} \&
  \bibinfo{author}{Hinton, G.~E.}
\newblock \bibinfo{title}{Imagenet classification with deep convolutional
  neural networks}.
\newblock In \emph{\bibinfo{booktitle}{Advances in neural information
  processing systems}}, \bibinfo{pages}{1097--1105} (\bibinfo{year}{2012}).

\bibitem{girshick2015rcnn}
\bibinfo{author}{Girshick, R.}
\newblock \bibinfo{title}{Fast r-cnn}.
\newblock In \emph{\bibinfo{booktitle}{Proceedings of the IEEE international
  conference on computer vision}}, \bibinfo{pages}{1440--1448}
  (\bibinfo{year}{2015}).

\bibitem{badrinarayanan2017segnet}
\bibinfo{author}{Badrinarayanan, V.}, \bibinfo{author}{Kendall, A.} \&
  \bibinfo{author}{Cipolla, R.}
\newblock \bibinfo{journal}{\bibinfo{title}{Segnet: A deep convolutional
  encoder-decoder architecture for image segmentation}}.
\newblock {\emph{\JournalTitle{IEEE transactions on pattern analysis and
  machine intelligence}}} \textbf{\bibinfo{volume}{39}},
  \bibinfo{pages}{2481--2495} (\bibinfo{year}{2017}).

\bibitem{zhang2017beyond}
\bibinfo{author}{Zhang, K.}, \bibinfo{author}{Zuo, W.}, \bibinfo{author}{Chen,
  Y.}, \bibinfo{author}{Meng, D.} \& \bibinfo{author}{Zhang, L.}
\newblock \bibinfo{journal}{\bibinfo{title}{Beyond a gaussian denoiser:
  Residual learning of deep cnn for image denoising}}.
\newblock {\emph{\JournalTitle{IEEE transactions on image processing}}}
  \textbf{\bibinfo{volume}{26}}, \bibinfo{pages}{3142--3155}
  (\bibinfo{year}{2017}).

\bibitem{VGG_Simonyan}
\bibinfo{author}{Simonyan, K.} \& \bibinfo{author}{Zisserman, A.}
\newblock \bibinfo{journal}{\bibinfo{title}{Very deep convolutional networks
  for large-scale image recognition}}.
\newblock {\emph{\JournalTitle{arXiv preprint arXiv:1409.1556}}}
  (\bibinfo{year}{2014}).

\bibitem{ResNet_He}
\bibinfo{author}{He, K.}, \bibinfo{author}{Zhang, X.}, \bibinfo{author}{Ren,
  S.} \& \bibinfo{author}{Sun, J.}
\newblock \bibinfo{title}{Deep residual learning for image recognition}.
\newblock In \emph{\bibinfo{booktitle}{Proceedings of the IEEE conference on
  computer vision and pattern recognition}}, \bibinfo{pages}{770--778}
  (\bibinfo{year}{2016}).

\bibitem{ResNeXt_Xie}
\bibinfo{author}{Xie, S.}, \bibinfo{author}{Girshick, R.},
  \bibinfo{author}{Doll{\'a}r, P.}, \bibinfo{author}{Tu, Z.} \&
  \bibinfo{author}{He, K.}
\newblock \bibinfo{title}{Aggregated residual transformations for deep neural
  networks}.
\newblock In \emph{\bibinfo{booktitle}{Proceedings of the IEEE conference on
  computer vision and pattern recognition}}, \bibinfo{pages}{1492--1500}
  (\bibinfo{year}{2017}).

\bibitem{szegedy2016inception_v3}
\bibinfo{author}{Szegedy, C.}, \bibinfo{author}{Vanhoucke, V.},
  \bibinfo{author}{Ioffe, S.}, \bibinfo{author}{Shlens, J.} \&
  \bibinfo{author}{Wojna, Z.}
\newblock \bibinfo{title}{Rethinking the inception architecture for computer
  vision}.
\newblock In \emph{\bibinfo{booktitle}{Proceedings of the IEEE conference on
  computer vision and pattern recognition}}, \bibinfo{pages}{2818--2826}
  (\bibinfo{year}{2016}).

\bibitem{DenseNet_Huang}
\bibinfo{author}{Huang, G.}, \bibinfo{author}{Liu, Z.}, \bibinfo{author}{Van
  Der~Maaten, L.} \& \bibinfo{author}{Weinberger, K.~Q.}
\newblock \bibinfo{title}{Densely connected convolutional networks}.
\newblock In \emph{\bibinfo{booktitle}{Proceedings of the IEEE conference on
  computer vision and pattern recognition}}, \bibinfo{pages}{4700--4708}
  (\bibinfo{year}{2017}).

\bibitem{krizhevsky2009learning}
\bibinfo{author}{Krizhevsky, A.}, \bibinfo{author}{Hinton, G.} \emph{et~al.}
\newblock \bibinfo{journal}{\bibinfo{title}{Learning multiple layers of
  features from tiny images}}.
\newblock {\emph{\JournalTitle{CoRR}}}  (\bibinfo{year}{2009}).

\bibitem{deng2009imagenet}
\bibinfo{author}{Deng, J.} \emph{et~al.}
\newblock \bibinfo{title}{Imagenet: A large-scale hierarchical image database}.
\newblock In \emph{\bibinfo{booktitle}{2009 IEEE conference on computer vision
  and pattern recognition}}, \bibinfo{pages}{248--255}
  (\bibinfo{organization}{Ieee}, \bibinfo{year}{2009}).

\bibitem{lecun1998gradient}
\bibinfo{author}{LeCun, Y.}, \bibinfo{author}{Bottou, L.},
  \bibinfo{author}{Bengio, Y.} \& \bibinfo{author}{Haffner, P.}
\newblock \bibinfo{journal}{\bibinfo{title}{Gradient-based learning applied to
  document recognition}}.
\newblock {\emph{\JournalTitle{Proceedings of the IEEE}}}
  \textbf{\bibinfo{volume}{86}}, \bibinfo{pages}{2278--2324}
  (\bibinfo{year}{1998}).

\bibitem{bycroft2018ukb}
\bibinfo{author}{Bycroft, C.} \emph{et~al.}
\newblock \bibinfo{journal}{\bibinfo{title}{The uk biobank resource with deep
  phenotyping and genomic data}}.
\newblock {\emph{\JournalTitle{Nature}}} \textbf{\bibinfo{volume}{562}},
  \bibinfo{pages}{203--209} (\bibinfo{year}{2018}).

\bibitem{van2013hcp}
\bibinfo{author}{Van~Essen, D.~C.} \emph{et~al.}
\newblock \bibinfo{journal}{\bibinfo{title}{The wu-minn human connectome
  project: an overview}}.
\newblock {\emph{\JournalTitle{Neuroimage}}} \textbf{\bibinfo{volume}{80}},
  \bibinfo{pages}{62--79} (\bibinfo{year}{2013}).

\bibitem{yang2017quicksilver}
\bibinfo{author}{Yang, X.}, \bibinfo{author}{Kwitt, R.},
  \bibinfo{author}{Styner, M.} \& \bibinfo{author}{Niethammer, M.}
\newblock \bibinfo{journal}{\bibinfo{title}{Quicksilver: Fast predictive image
  registration--a deep learning approach}}.
\newblock {\emph{\JournalTitle{NeuroImage}}} \textbf{\bibinfo{volume}{158}},
  \bibinfo{pages}{378--396} (\bibinfo{year}{2017}).

\bibitem{havaei2017braintumor}
\bibinfo{author}{Havaei, M.} \emph{et~al.}
\newblock \bibinfo{journal}{\bibinfo{title}{Brain tumor segmentation with deep
  neural networks}}.
\newblock {\emph{\JournalTitle{Medical image analysis}}}
  \textbf{\bibinfo{volume}{35}}, \bibinfo{pages}{18--31}
  (\bibinfo{year}{2017}).

\bibitem{wen2020}
\bibinfo{author}{Wen, J.} \emph{et~al.}
\newblock \bibinfo{journal}{\bibinfo{title}{Convolutional neural networks for
  classification of alzheimer's disease: Overview and reproducible
  evaluation}}.
\newblock {\emph{\JournalTitle{Medical Image Analysis}}}
  \bibinfo{pages}{101694} (\bibinfo{year}{2020}).

\bibitem{plis2014deep}
\bibinfo{author}{Plis, S.~M.} \emph{et~al.}
\newblock \bibinfo{journal}{\bibinfo{title}{Deep learning for neuroimaging: a
  validation study}}.
\newblock {\emph{\JournalTitle{Frontiers in neuroscience}}}
  \textbf{\bibinfo{volume}{8}}, \bibinfo{pages}{229} (\bibinfo{year}{2014}).

\bibitem{sujit2019automated}
\bibinfo{author}{Sujit, S.~J.}, \bibinfo{author}{Coronado, I.},
  \bibinfo{author}{Kamali, A.}, \bibinfo{author}{Narayana, P.~A.} \&
  \bibinfo{author}{Gabr, R.~E.}
\newblock \bibinfo{journal}{\bibinfo{title}{Automated image quality evaluation
  of structural brain mri using an ensemble of deep learning networks}}.
\newblock {\emph{\JournalTitle{Journal of Magnetic Resonance Imaging}}}
  \textbf{\bibinfo{volume}{50}}, \bibinfo{pages}{1260--1267}
  (\bibinfo{year}{2019}).

\bibitem{shahamat2020brain}
\bibinfo{author}{Shahamat, H.} \& \bibinfo{author}{Abadeh, M.~S.}
\newblock \bibinfo{journal}{\bibinfo{title}{Brain mri analysis using a deep
  learning based volutionary approach}}.
\newblock {\emph{\JournalTitle{Neural Networks}}}  (\bibinfo{year}{2020}).

\bibitem{schulz2020}
\bibinfo{author}{Schulz, M.-A.} \emph{et~al.}
\newblock \bibinfo{journal}{\bibinfo{title}{Different scaling of linear models
  and deep learning in ukbiobank brain images versus machine-learning
  datasets}}.
\newblock {\emph{\JournalTitle{Nature communications}}}
  \textbf{\bibinfo{volume}{11}}, \bibinfo{pages}{1--15} (\bibinfo{year}{2020}).

\bibitem{Jonsson2019}
\bibinfo{author}{Jonsson, B.~A.} \emph{et~al.}
\newblock \bibinfo{journal}{\bibinfo{title}{Brain age prediction using deep
  learning uncovers associated sequence variants}}.
\newblock {\emph{\JournalTitle{Nature Communications}}}
  \textbf{\bibinfo{volume}{10}}, \doiprefix\url{10.1038/s41467-019-13163-9}
  (\bibinfo{year}{2019}).

\bibitem{guo2017calibration}
\bibinfo{author}{Guo, C.}, \bibinfo{author}{Pleiss, G.}, \bibinfo{author}{Sun,
  Y.} \& \bibinfo{author}{Weinberger, K.~Q.}
\newblock \bibinfo{title}{On calibration of modern neural networks}.
\newblock In \emph{\bibinfo{booktitle}{International Conference on Machine
  Learning}}, \bibinfo{pages}{1321--1330} (\bibinfo{year}{2017}).

\bibitem{gal2016phd}
\bibinfo{author}{Gal, Y.}
\newblock \bibinfo{journal}{\bibinfo{title}{Uncertainty in deep learning}}.
\newblock {\emph{\JournalTitle{University of Cambridge}}}
  \textbf{\bibinfo{volume}{1}}, \bibinfo{pages}{3} (\bibinfo{year}{2016}).

\bibitem{lakshminarayanan2017}
\bibinfo{author}{Lakshminarayanan, B.}, \bibinfo{author}{Pritzel, A.} \&
  \bibinfo{author}{Blundell, C.}
\newblock \bibinfo{title}{Simple and scalable predictive uncertainty estimation
  using deep ensembles}.
\newblock In \emph{\bibinfo{booktitle}{Advances in neural information
  processing systems}}, \bibinfo{pages}{6402--6413} (\bibinfo{year}{2017}).

\bibitem{gustafsson2020bench}
\bibinfo{author}{Gustafsson, F.~K.}, \bibinfo{author}{Danelljan, M.} \&
  \bibinfo{author}{Schon, T.~B.}
\newblock \bibinfo{title}{Evaluating scalable bayesian deep learning methods
  for robust computer vision}.
\newblock In \emph{\bibinfo{booktitle}{Proceedings of the IEEE/CVF Conference
  on Computer Vision and Pattern Recognition Workshops}},
  \bibinfo{pages}{318--319} (\bibinfo{year}{2020}).

\bibitem{ashburner2007dartel}
\bibinfo{author}{Ashburner, J.}
\newblock \bibinfo{journal}{\bibinfo{title}{A fast diffeomorphic image
  registration algorithm}}.
\newblock {\emph{\JournalTitle{Neuroimage}}} \textbf{\bibinfo{volume}{38}},
  \bibinfo{pages}{95--113} (\bibinfo{year}{2007}).

\bibitem{peng2021}
\bibinfo{author}{Peng, H.}, \bibinfo{author}{Gong, W.},
  \bibinfo{author}{Beckmann, C.~F.}, \bibinfo{author}{Vedaldi, A.} \&
  \bibinfo{author}{Smith, S.~M.}
\newblock \bibinfo{journal}{\bibinfo{title}{Accurate brain age prediction with
  lightweight deep neural networks}}.
\newblock {\emph{\JournalTitle{Medical Image Analysis}}}
  \textbf{\bibinfo{volume}{68}}, \bibinfo{pages}{101871}
  (\bibinfo{year}{2021}).

\bibitem{sturmfels2018domain}
\bibinfo{author}{Sturmfels, P.} \emph{et~al.}
\newblock \bibinfo{title}{A domain guided cnn architecture for predicting age
  from structural brain images}.
\newblock In \emph{\bibinfo{booktitle}{Machine Learning for Healthcare
  Conference}}, \bibinfo{pages}{295--311} (\bibinfo{year}{2018}).

\bibitem{cole2017predicting}
\bibinfo{author}{Cole, J.~H.} \emph{et~al.}
\newblock \bibinfo{journal}{\bibinfo{title}{Predicting brain age with deep
  learning from raw imaging data results in a reliable and heritable
  biomarker}}.
\newblock {\emph{\JournalTitle{NeuroImage}}} \textbf{\bibinfo{volume}{163}},
  \bibinfo{pages}{115--124} (\bibinfo{year}{2017}).

\bibitem{ueda2019age}
\bibinfo{author}{Ueda, M.} \emph{et~al.}
\newblock \bibinfo{title}{An age estimation method using 3d-cnn from brain mri
  images}.
\newblock In \emph{\bibinfo{booktitle}{2019 IEEE 16th International Symposium
  on Biomedical Imaging (ISBI 2019)}}, \bibinfo{pages}{380--383}
  (\bibinfo{organization}{IEEE}, \bibinfo{year}{2019}).

\bibitem{armanious2020age}
\bibinfo{author}{Armanious, K.} \emph{et~al.}
\newblock \bibinfo{journal}{\bibinfo{title}{Age-net: An mri-based iterative
  framework for biological age estimation}}.
\newblock {\emph{\JournalTitle{arXiv preprint arXiv:2009.10765}}}
  (\bibinfo{year}{2020}).

\bibitem{varatharajah2018}
\bibinfo{author}{Varatharajah, Y.} \emph{et~al.}
\newblock \bibinfo{journal}{\bibinfo{title}{Predicting brain age using
  structural neuroimaging and deep learning}}.
\newblock {\emph{\JournalTitle{bioRxiv}}} \bibinfo{pages}{497925}
  (\bibinfo{year}{2018}).

\bibitem{bashyam2020mri}
\bibinfo{author}{Bashyam, V.~M.} \emph{et~al.}
\newblock \bibinfo{journal}{\bibinfo{title}{Mri signatures of brain age and
  disease over the lifespan based on a deep brain network and 14 468
  individuals worldwide}}.
\newblock {\emph{\JournalTitle{Brain}}} \textbf{\bibinfo{volume}{143}},
  \bibinfo{pages}{2312--2324} (\bibinfo{year}{2020}).

\bibitem{li2017ad}
\bibinfo{author}{Li, F.}, \bibinfo{author}{Cheng, D.} \& \bibinfo{author}{Liu,
  M.}
\newblock \bibinfo{title}{Alzheimer's disease classification based on
  combination of multi-model convolutional networks}.
\newblock In \emph{\bibinfo{booktitle}{2017 IEEE International Conference on
  Imaging Systems and Techniques (IST)}}, \bibinfo{pages}{1--5}
  (\bibinfo{organization}{IEEE}, \bibinfo{year}{2017}).

\bibitem{backstrom2018efficient}
\bibinfo{author}{B{\"a}ckstr{\"o}m, K.}, \bibinfo{author}{Nazari, M.},
  \bibinfo{author}{Gu, I. Y.-H.} \& \bibinfo{author}{Jakola, A.~S.}
\newblock \bibinfo{title}{An efficient 3d deep convolutional network for
  alzheimer's disease diagnosis using mr images}.
\newblock In \emph{\bibinfo{booktitle}{2018 IEEE 15th International Symposium
  on Biomedical Imaging (ISBI 2018)}}, \bibinfo{pages}{149--153}
  (\bibinfo{organization}{IEEE}, \bibinfo{year}{2018}).

\bibitem{korolev2017}
\bibinfo{author}{Korolev, S.}, \bibinfo{author}{Safiullin, A.},
  \bibinfo{author}{Belyaev, M.} \& \bibinfo{author}{Dodonova, Y.}
\newblock \bibinfo{title}{Residual and plain convolutional neural networks for
  3d brain mri classification}.
\newblock In \emph{\bibinfo{booktitle}{2017 IEEE 14th International Symposium
  on Biomedical Imaging (ISBI 2017)}}, \bibinfo{pages}{835--838}
  (\bibinfo{organization}{IEEE}, \bibinfo{year}{2017}).

\bibitem{hosseini2018}
\bibinfo{author}{Hosseini-Asl, E.} \emph{et~al.}
\newblock \bibinfo{journal}{\bibinfo{title}{Alzheimer's disease diagnostics by
  a 3d deeply supervised adaptable convolutional network.}}
\newblock {\emph{\JournalTitle{Frontiers in bioscience (Landmark edition)}}}
  \textbf{\bibinfo{volume}{23}}, \bibinfo{pages}{584} (\bibinfo{year}{2018}).

\bibitem{shmulev2018}
\bibinfo{author}{Shmulev, Y.}, \bibinfo{author}{Belyaev, M.},
  \bibinfo{author}{Initiative, A. D.~N.} \emph{et~al.}
\newblock \bibinfo{title}{Predicting conversion of mild cognitive impairments
  to alzheimer’s disease and exploring impact of neuroimaging}.
\newblock In \emph{\bibinfo{booktitle}{Graphs in Biomedical Image Analysis and
  Integrating Medical Imaging and Non-Imaging Modalities}},
  \bibinfo{pages}{83--91} (\bibinfo{publisher}{Springer},
  \bibinfo{year}{2018}).

\bibitem{abrol2020}
\bibinfo{author}{Abrol, A.} \emph{et~al.}
\newblock \bibinfo{journal}{\bibinfo{title}{Deep residual learning for
  neuroimaging: An application to predict progression to alzheimer’s
  disease}}.
\newblock {\emph{\JournalTitle{Journal of Neuroscience Methods}}}
  \bibinfo{pages}{108701} (\bibinfo{year}{2020}).

\bibitem{senanayake2018}
\bibinfo{author}{Senanayake, U.}, \bibinfo{author}{Sowmya, A.} \&
  \bibinfo{author}{Dawes, L.}
\newblock \bibinfo{title}{Deep fusion pipeline for mild cognitive impairment
  diagnosis}.
\newblock In \emph{\bibinfo{booktitle}{2018 IEEE 15th International Symposium
  on Biomedical Imaging (ISBI 2018)}}, \bibinfo{pages}{1394--1997}
  (\bibinfo{organization}{IEEE}, \bibinfo{year}{2018}).

\bibitem{spasov2019}
\bibinfo{author}{Spasov, S.} \emph{et~al.}
\newblock \bibinfo{journal}{\bibinfo{title}{A parameter-efficient deep learning
  approach to predict conversion from mild cognitive impairment to alzheimer's
  disease}}.
\newblock {\emph{\JournalTitle{Neuroimage}}} \textbf{\bibinfo{volume}{189}},
  \bibinfo{pages}{276--287} (\bibinfo{year}{2019}).

\bibitem{wang2019}
\bibinfo{author}{Wang, H.} \emph{et~al.}
\newblock \bibinfo{journal}{\bibinfo{title}{Ensemble of 3d densely connected
  convolutional network for diagnosis of mild cognitive impairment and
  alzheimer’s disease}}.
\newblock {\emph{\JournalTitle{Neurocomputing}}}
  \textbf{\bibinfo{volume}{333}}, \bibinfo{pages}{145--156}
  (\bibinfo{year}{2019}).

\bibitem{hu2020}
\bibinfo{author}{Hu, M.}, \bibinfo{author}{Sim, K.}, \bibinfo{author}{Zhou,
  J.~H.}, \bibinfo{author}{Jiang, X.} \& \bibinfo{author}{Guan, C.}
\newblock \bibinfo{journal}{\bibinfo{title}{Brain mri-based 3d convolutional
  neural networks for classification of schizophrenia and controls}}.
\newblock {\emph{\JournalTitle{arXiv preprint arXiv:2003.08818}}}
  (\bibinfo{year}{2020}).

\bibitem{oh2020}
\bibinfo{author}{Oh, J.}, \bibinfo{author}{Oh, B.-L.}, \bibinfo{author}{Lee,
  K.-U.}, \bibinfo{author}{Chae, J.-H.} \& \bibinfo{author}{Yun, K.}
\newblock \bibinfo{journal}{\bibinfo{title}{Identifying schizophrenia using
  structural mri with a deep learning algorithm}}.
\newblock {\emph{\JournalTitle{Frontiers in psychiatry}}}
  \textbf{\bibinfo{volume}{11}}, \bibinfo{pages}{16} (\bibinfo{year}{2020}).

\bibitem{iandola2016squeezenet}
\bibinfo{author}{Iandola, F.~N.} \emph{et~al.}
\newblock \bibinfo{journal}{\bibinfo{title}{Squeezenet: Alexnet-level accuracy
  with 50x fewer parameters and< 0.5 mb model size}}.
\newblock {\emph{\JournalTitle{arXiv preprint arXiv:1602.07360}}}
  (\bibinfo{year}{2016}).

\bibitem{gould2014svm_scz}
\bibinfo{author}{Gould, I.~C.} \emph{et~al.}
\newblock \bibinfo{journal}{\bibinfo{title}{Multivariate neuroanatomical
  classification of cognitive subtypes in schizophrenia: a support vector
  machine learning approach}}.
\newblock {\emph{\JournalTitle{NeuroImage: Clinical}}}
  \textbf{\bibinfo{volume}{6}}, \bibinfo{pages}{229--236}
  (\bibinfo{year}{2014}).

\bibitem{depierrefeu2018identifying}
\bibinfo{author}{de~Pierrefeu, A.} \emph{et~al.}
\newblock \bibinfo{journal}{\bibinfo{title}{Identifying a neuroanatomical
  signature of schizophrenia, reproducible across sites and stages, using
  machine learning with structured sparsity}}.
\newblock {\emph{\JournalTitle{Acta Psychiatrica Scandinavica}}}
  \textbf{\bibinfo{volume}{138}}, \bibinfo{pages}{571--580}
  (\bibinfo{year}{2018}).

\bibitem{xiao2019svm}
\bibinfo{author}{Xiao, Y.} \emph{et~al.}
\newblock \bibinfo{journal}{\bibinfo{title}{Support vector machine-based
  classification of first episode drug-na{\"\i}ve schizophrenia patients and
  healthy controls using structural mri}}.
\newblock {\emph{\JournalTitle{Schizophrenia Research}}}
  \textbf{\bibinfo{volume}{214}}, \bibinfo{pages}{11--17}
  (\bibinfo{year}{2019}).

\bibitem{vieira2020using}
\bibinfo{author}{Vieira, S.} \emph{et~al.}
\newblock \bibinfo{journal}{\bibinfo{title}{Using machine learning and
  structural neuroimaging to detect first episode psychosis: reconsidering the
  evidence}}.
\newblock {\emph{\JournalTitle{Schizophrenia bulletin}}}
  \textbf{\bibinfo{volume}{46}}, \bibinfo{pages}{17--26}
  (\bibinfo{year}{2020}).

\bibitem{pinaya2016dbn}
\bibinfo{author}{Pinaya, W.~H.} \emph{et~al.}
\newblock \bibinfo{journal}{\bibinfo{title}{Using deep belief network modelling
  to characterize differences in brain morphometry in schizophrenia}}.
\newblock {\emph{\JournalTitle{Scientific reports}}}
  \textbf{\bibinfo{volume}{6}}, \bibinfo{pages}{38897} (\bibinfo{year}{2016}).

\bibitem{latha2019}
\bibinfo{author}{Latha, M.} \& \bibinfo{author}{Kavitha, G.}
\newblock \bibinfo{journal}{\bibinfo{title}{Detection of schizophrenia in brain
  mr images based on segmented ventricle region and deep belief networks}}.
\newblock {\emph{\JournalTitle{Neural Computing and Applications}}}
  \textbf{\bibinfo{volume}{31}}, \bibinfo{pages}{5195--5206}
  (\bibinfo{year}{2019}).

\bibitem{pinaya2019}
\bibinfo{author}{Pinaya, W.~H.}, \bibinfo{author}{Mechelli, A.} \&
  \bibinfo{author}{Sato, J.~R.}
\newblock \bibinfo{journal}{\bibinfo{title}{Using deep autoencoders to identify
  abnormal brain structural patterns in neuropsychiatric disorders: A
  large-scale multi-sample study}}.
\newblock {\emph{\JournalTitle{Human brain mapping}}}
  \textbf{\bibinfo{volume}{40}}, \bibinfo{pages}{944--954}
  (\bibinfo{year}{2019}).

\bibitem{abrol2021deep}
\bibinfo{author}{Abrol, A.} \emph{et~al.}
\newblock \bibinfo{journal}{\bibinfo{title}{Deep learning encodes robust
  discriminative neuroimaging representations to outperform standard machine
  learning}}.
\newblock {\emph{\JournalTitle{Nature communications}}}
  \textbf{\bibinfo{volume}{12}}, \bibinfo{pages}{1--17} (\bibinfo{year}{2021}).

\bibitem{goodfellow2016deep}
\bibinfo{author}{Goodfellow, I.}, \bibinfo{author}{Bengio, Y.},
  \bibinfo{author}{Courville, A.} \& \bibinfo{author}{Bengio, Y.}
\newblock \emph{\bibinfo{title}{Deep learning}}, vol.~\bibinfo{volume}{1}
  (\bibinfo{publisher}{MIT press Cambridge}, \bibinfo{year}{2016}).

\bibitem{lecun2015deep}
\bibinfo{author}{LeCun, Y.}, \bibinfo{author}{Bengio, Y.} \&
  \bibinfo{author}{Hinton, G.}
\newblock \bibinfo{journal}{\bibinfo{title}{Deep learning}}.
\newblock {\emph{\JournalTitle{nature}}} \textbf{\bibinfo{volume}{521}},
  \bibinfo{pages}{436--444} (\bibinfo{year}{2015}).

\bibitem{tamminga2014bipolar}
\bibinfo{author}{Tamminga, C.~A.} \emph{et~al.}
\newblock \bibinfo{journal}{\bibinfo{title}{Bipolar and schizophrenia network
  for intermediate phenotypes: outcomes across the psychosis continuum}}.
\newblock {\emph{\JournalTitle{Schizophrenia bulletin}}}
  \textbf{\bibinfo{volume}{40}}, \bibinfo{pages}{S131--S137}
  (\bibinfo{year}{2014}).

\bibitem{hozer2020biobd}
\bibinfo{author}{Hozer, F.} \emph{et~al.}
\newblock \bibinfo{journal}{\bibinfo{title}{Lithium prevents grey matter
  atrophy in patients with bipolar disorder: an international multicenter
  study}}.
\newblock {\emph{\JournalTitle{Psychological medicine}}} \bibinfo{pages}{1--10}
  (\bibinfo{year}{2020}).

\bibitem{gaser2016cat12vbm}
\bibinfo{author}{Gaser, C.} \& \bibinfo{author}{Dahnke, R.}
\newblock \bibinfo{journal}{\bibinfo{title}{Cat-a computational anatomy toolbox
  for the analysis of structural mri data}}.
\newblock {\emph{\JournalTitle{HBM}}} \textbf{\bibinfo{volume}{2016}},
  \bibinfo{pages}{336--348} (\bibinfo{year}{2016}).

\bibitem{tustison2010n4itk}
\bibinfo{author}{Tustison, N.~J.} \emph{et~al.}
\newblock \bibinfo{journal}{\bibinfo{title}{N4itk: improved n3 bias
  correction}}.
\newblock {\emph{\JournalTitle{IEEE transactions on medical imaging}}}
  \textbf{\bibinfo{volume}{29}}, \bibinfo{pages}{1310--1320}
  (\bibinfo{year}{2010}).

\bibitem{avants2009ants}
\bibinfo{author}{Avants, B.~B.}, \bibinfo{author}{Tustison, N.} \&
  \bibinfo{author}{Song, G.}
\newblock \bibinfo{journal}{\bibinfo{title}{Advanced normalization tools
  (ants)}}.
\newblock {\emph{\JournalTitle{Insight j}}} \textbf{\bibinfo{volume}{2}},
  \bibinfo{pages}{1--35} (\bibinfo{year}{2009}).

\bibitem{jenkinson2005bet2}
\bibinfo{author}{Jenkinson, M.}, \bibinfo{author}{Pechaud, M.},
  \bibinfo{author}{Smith, S.} \emph{et~al.}
\newblock \bibinfo{title}{Bet2: Mr-based estimation of brain, skull and scalp
  surfaces}.
\newblock In \emph{\bibinfo{booktitle}{Eleventh annual meeting of the
  organization for human brain mapping}}, vol.~\bibinfo{volume}{17},
  \bibinfo{pages}{167} (\bibinfo{organization}{Toronto.},
  \bibinfo{year}{2005}).

\bibitem{jenkinson2001flirt}
\bibinfo{author}{Jenkinson, M.} \& \bibinfo{author}{Smith, S.}
\newblock \bibinfo{journal}{\bibinfo{title}{A global optimisation method for
  robust affine registration of brain images}}.
\newblock {\emph{\JournalTitle{Medical image analysis}}}
  \textbf{\bibinfo{volume}{5}}, \bibinfo{pages}{143--156}
  (\bibinfo{year}{2001}).

\bibitem{raghu2017svcca}
\bibinfo{author}{Raghu, M.}, \bibinfo{author}{Gilmer, J.},
  \bibinfo{author}{Yosinski, J.} \& \bibinfo{author}{Sohl-Dickstein, J.}
\newblock \bibinfo{title}{Svcca: Singular vector canonical correlation analysis
  for deep learning dynamics and interpretability}.
\newblock In \emph{\bibinfo{booktitle}{Advances in Neural Information
  Processing Systems}}, \bibinfo{pages}{6076--6085} (\bibinfo{year}{2017}).

\bibitem{arlot:hal-00407906}
\bibinfo{author}{Arlot, S.} \& \bibinfo{author}{Celisse, A.}
\newblock \bibinfo{journal}{\bibinfo{title}{{A survey of cross-validation
  procedures for model selection}}}.
\newblock {\emph{\JournalTitle{{Statistics Surveys}}}}
  \textbf{\bibinfo{volume}{4}}, \bibinfo{pages}{40--79} (\bibinfo{year}{2010}).

\bibitem{shaw2019_motion_artefact}
\bibinfo{author}{Shaw, R.}, \bibinfo{author}{Sudre, C.~H.},
  \bibinfo{author}{Ourselin, S.} \& \bibinfo{author}{Cardoso, M.~J.}
\newblock \bibinfo{title}{Mri k-space motion artefact augmentation: Model
  robustness and task-specific uncertainty.}
\newblock In \emph{\bibinfo{booktitle}{MIDL}}, \bibinfo{pages}{427--436}
  (\bibinfo{year}{2019}).

\bibitem{zhuo2006_MRI_artefacts}
\bibinfo{author}{Zhuo, J.} \& \bibinfo{author}{Gullapalli, R.~P.}
\newblock \bibinfo{journal}{\bibinfo{title}{Mr artifacts, safety, and quality
  control}}.
\newblock {\emph{\JournalTitle{Radiographics}}} \textbf{\bibinfo{volume}{26}},
  \bibinfo{pages}{275--297} (\bibinfo{year}{2006}).

\bibitem{van1999}
\bibinfo{author}{Van~Leemput, K.}, \bibinfo{author}{Maes, F.},
  \bibinfo{author}{Vandermeulen, D.} \& \bibinfo{author}{Suetens, P.}
\newblock \bibinfo{journal}{\bibinfo{title}{Automated model-based tissue
  classification of mr images of the brain}}.
\newblock {\emph{\JournalTitle{IEEE transactions on medical imaging}}}
  \textbf{\bibinfo{volume}{18}}, \bibinfo{pages}{897--908}
  (\bibinfo{year}{1999}).

\bibitem{chen2019self_supervision}
\bibinfo{author}{Chen, L.} \emph{et~al.}
\newblock \bibinfo{journal}{\bibinfo{title}{Self-supervised learning for
  medical image analysis using image context restoration}}.
\newblock {\emph{\JournalTitle{Medical image analysis}}}
  \textbf{\bibinfo{volume}{58}}, \bibinfo{pages}{101539}
  (\bibinfo{year}{2019}).

\bibitem{kendall2017uncertainties}
\bibinfo{author}{Kendall, A.} \& \bibinfo{author}{Gal, Y.}
\newblock \bibinfo{journal}{\bibinfo{title}{What uncertainties do we need in
  bayesian deep learning for computer vision?}}
\newblock {\emph{\JournalTitle{arXiv preprint arXiv:1703.04977}}}
  (\bibinfo{year}{2017}).

\bibitem{gal2016dropout}
\bibinfo{author}{Gal, Y.} \& \bibinfo{author}{Ghahramani, Z.}
\newblock \bibinfo{title}{Dropout as a bayesian approximation: Representing
  model uncertainty in deep learning}.
\newblock In \emph{\bibinfo{booktitle}{international conference on machine
  learning}}, \bibinfo{pages}{1050--1059} (\bibinfo{organization}{PMLR},
  \bibinfo{year}{2016}).

\bibitem{filos2019systematic}
\bibinfo{author}{Filos, A.} \emph{et~al.}
\newblock \bibinfo{journal}{\bibinfo{title}{A systematic comparison of bayesian
  deep learning robustness in diabetic retinopathy tasks}}.
\newblock {\emph{\JournalTitle{arXiv preprint arXiv:1912.10481}}}
  (\bibinfo{year}{2019}).

\bibitem{leibig2017leveraging}
\bibinfo{author}{Leibig, C.}, \bibinfo{author}{Allken, V.},
  \bibinfo{author}{Ayhan, M.~S.}, \bibinfo{author}{Berens, P.} \&
  \bibinfo{author}{Wahl, S.}
\newblock \bibinfo{journal}{\bibinfo{title}{Leveraging uncertainty information
  from deep neural networks for disease detection}}.
\newblock {\emph{\JournalTitle{Scientific reports}}}
  \textbf{\bibinfo{volume}{7}}, \bibinfo{pages}{1--14} (\bibinfo{year}{2017}).

\bibitem{gal2017concrete_dropout}
\bibinfo{author}{Gal, Y.}, \bibinfo{author}{Hron, J.} \&
  \bibinfo{author}{Kendall, A.}
\newblock \bibinfo{title}{Concrete dropout}.
\newblock In \emph{\bibinfo{booktitle}{Advances in neural information
  processing systems}}, \bibinfo{pages}{3581--3590} (\bibinfo{year}{2017}).

\bibitem{kingma2014adam}
\bibinfo{author}{Kingma, D.~P.} \& \bibinfo{author}{Ba, J.}
\newblock \bibinfo{journal}{\bibinfo{title}{Adam: A method for stochastic
  optimization}}.
\newblock {\emph{\JournalTitle{arXiv preprint arXiv:1412.6980}}}
  (\bibinfo{year}{2014}).

\bibitem{paszke2019pytorch}
\bibinfo{author}{Paszke, A.} \emph{et~al.}
\newblock \bibinfo{journal}{\bibinfo{title}{Pytorch: An imperative style,
  high-performance deep learning library}}.
\newblock {\emph{\JournalTitle{arXiv preprint arXiv:1912.01703}}}
  (\bibinfo{year}{2019}).

\bibitem{pedregosa2011scikit}
\bibinfo{author}{Pedregosa, F.} \emph{et~al.}
\newblock \bibinfo{journal}{\bibinfo{title}{Scikit-learn: Machine learning in
  python}}.
\newblock {\emph{\JournalTitle{the Journal of machine Learning research}}}
  \textbf{\bibinfo{volume}{12}}, \bibinfo{pages}{2825--2830}
  (\bibinfo{year}{2011}).

\bibitem{wachinger2021detect}
\bibinfo{author}{Wachinger, C.}, \bibinfo{author}{Rieckmann, A.},
  \bibinfo{author}{P{\"o}lsterl, S.}, \bibinfo{author}{Initiative, A. D.~N.}
  \emph{et~al.}
\newblock \bibinfo{journal}{\bibinfo{title}{Detect and correct bias in
  multi-site neuroimaging datasets}}.
\newblock {\emph{\JournalTitle{Medical Image Analysis}}}
  \textbf{\bibinfo{volume}{67}}, \bibinfo{pages}{101879}
  (\bibinfo{year}{2021}).

\bibitem{glocker2019machine}
\bibinfo{author}{Glocker, B.}, \bibinfo{author}{Robinson, R.},
  \bibinfo{author}{Castro, D.~C.}, \bibinfo{author}{Dou, Q.} \&
  \bibinfo{author}{Konukoglu, E.}
\newblock \bibinfo{journal}{\bibinfo{title}{Machine learning with multi-site
  imaging data: An empirical study on the impact of scanner effects}}.
\newblock {\emph{\JournalTitle{arXiv preprint arXiv:1910.04597}}}
  (\bibinfo{year}{2019}).

\bibitem{tartaglione2021end}
\bibinfo{author}{Tartaglione, E.}, \bibinfo{author}{Barbano, C.~A.} \&
  \bibinfo{author}{Grangetto, M.}
\newblock \bibinfo{journal}{\bibinfo{title}{End: Entangling and disentangling
  deep representations for bias correction}}.
\newblock {\emph{\JournalTitle{arXiv preprint arXiv:2103.02023}}}
  (\bibinfo{year}{2021}).

\bibitem{torralba2011unbiased}
\bibinfo{author}{Torralba, A.} \& \bibinfo{author}{Efros, A.~A.}
\newblock \bibinfo{title}{Unbiased look at dataset bias}.
\newblock In \emph{\bibinfo{booktitle}{CVPR 2011}}, \bibinfo{pages}{1521--1528}
  (\bibinfo{organization}{IEEE}, \bibinfo{year}{2011}).

\bibitem{kim2019learning}
\bibinfo{author}{Kim, B.}, \bibinfo{author}{Kim, H.}, \bibinfo{author}{Kim,
  K.}, \bibinfo{author}{Kim, S.} \& \bibinfo{author}{Kim, J.}
\newblock \bibinfo{title}{Learning not to learn: Training deep neural networks
  with biased data}.
\newblock In \emph{\bibinfo{booktitle}{Proceedings of the IEEE/CVF Conference
  on Computer Vision and Pattern Recognition}}, \bibinfo{pages}{9012--9020}
  (\bibinfo{year}{2019}).

\bibitem{hendricks2018women}
\bibinfo{author}{Hendricks, L.~A.}, \bibinfo{author}{Burns, K.},
  \bibinfo{author}{Saenko, K.}, \bibinfo{author}{Darrell, T.} \&
  \bibinfo{author}{Rohrbach, A.}
\newblock \bibinfo{title}{Women also snowboard: Overcoming bias in captioning
  models}.
\newblock In \emph{\bibinfo{booktitle}{Proceedings of the European Conference
  on Computer Vision (ECCV)}}, \bibinfo{pages}{771--787}
  (\bibinfo{year}{2018}).

\bibitem{perez2020torchio}
\bibinfo{author}{P{\'e}rez-Garc{\'\i}a, F.}, \bibinfo{author}{Sparks, R.} \&
  \bibinfo{author}{Ourselin, S.}
\newblock \bibinfo{journal}{\bibinfo{title}{Torchio: a python library for
  efficient loading, preprocessing, augmentation and patch-based sampling of
  medical images in deep learning}}.
\newblock {\emph{\JournalTitle{arXiv preprint arXiv:2003.04696}}}
  (\bibinfo{year}{2020}).

\bibitem{srivastava2014dropout}
\bibinfo{author}{Srivastava, N.}, \bibinfo{author}{Hinton, G.},
  \bibinfo{author}{Krizhevsky, A.}, \bibinfo{author}{Sutskever, I.} \&
  \bibinfo{author}{Salakhutdinov, R.}
\newblock \bibinfo{journal}{\bibinfo{title}{Dropout: a simple way to prevent
  neural networks from overfitting}}.
\newblock {\emph{\JournalTitle{The journal of machine learning research}}}
  \textbf{\bibinfo{volume}{15}}, \bibinfo{pages}{1929--1958}
  (\bibinfo{year}{2014}).

\end{thebibliography}

\section*{Supplementary Material}

\subsection*{Training and Test Split}

\begin{table}[h!]
        \centering
        \resizebox{.48\textwidth}{!}{  
        \begin{tabular}{|c|c|c|c|}
            \hline
            \rowcolor{Gray}
            \textbf{Task} & \textbf{Training Set}  & \textbf{Test Set} \\
            \hline
            Age & BHB-10K & BSNIP (only HC) \\
            \hline 
            Sex & BHB-10K & BSNIP (only HC) \\
            \hline
            Dx & SCHIZCONNECT-VIP & BSNIP \\
            \hline
        \end{tabular}
        }
    \captionof{table}{Training and Test Split used throughout this study for the 3 different target tasks. }
    \label{training_test_split_supp}
\end{table}

\subsection*{Impact of Spatial Resolution with Quasi-Raw MR Images}

    As we wanted to make sure that down-sampling the quasi-raw images from $1mm^3$ to $1.5mm^3$ isotropic had no negative impact on the final performance, we performed 3 experiments with ResNet18 and $N=500$ training samples with the same experimental design as in section \ref{cnn_perf_N500}. Results show no drop in performance for sex and Dx predictions and even an improvement for age regression. We believe it could be partly due to the noise in the data at such a resolution.
    \begin{table}[h]
        \centering
        \resizebox{.48\textwidth}{!}{ 
        \begin{tabular}{|c|c|c|c|c|}
            \rowcolor{Gray}
             \multicolumn{1}{c}{\textbf{Spatial Resolution}} & \multicolumn{2}{c}{\textbf{Age}} & \multicolumn{1}{c}{\textbf{Sex}} & \multicolumn{1}{c}{\textbf{Dx}}  \\
             \hline
                                & MAE & RMSE & AUC & AUC \\
            \cline{2-5}
            $1.5mm^3$ & $7.91 \pm 0.42$ &  $9.82 \pm 0.55$ & $0.84\pm 0.01$ & $0.66\pm 0.01$ \\
            $1mm^3$ & $12.5 \pm 1.6$ & $11.9 \pm 1.22$ & $0.84 \pm 0.02$ & $0.66 \pm 0.02$\\
            \hline  
        \end{tabular}
        }
        \caption{ResNet18 performance as we change the spatial resolution of the input images from $1mm^3$ (size $182\times218\times182$) to $1.5mm^3$ isotropic (size $121\times 145\times 121$). It mainly affects the computational burden (higher when using the $1mm^3$ resolution) and it even improves the results for age prediction.}
        \label{tab:spatial_res_impact}
    \end{table}

\subsection*{Introduction of tiny-DenseNet}

\begin{figure*}[h!] 
        \centering
        \resizebox{\textwidth}{!}{
        \subfloat[DenseNet121]{%
            \includegraphics[width=0.5\textwidth]{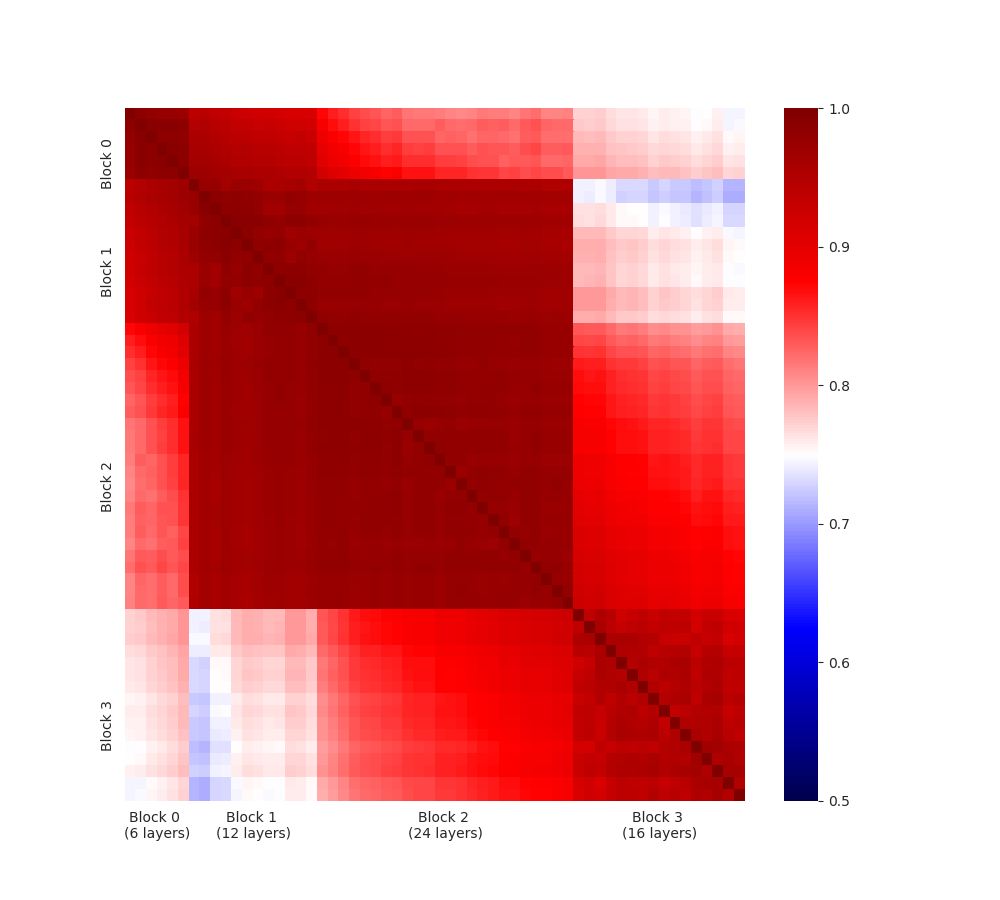}%
            \label{fig:densenet121_svcca}%
        }%
        \hfill%
        \subfloat[tiny-DenseNet]{%
            \includegraphics[width=0.5\textwidth]{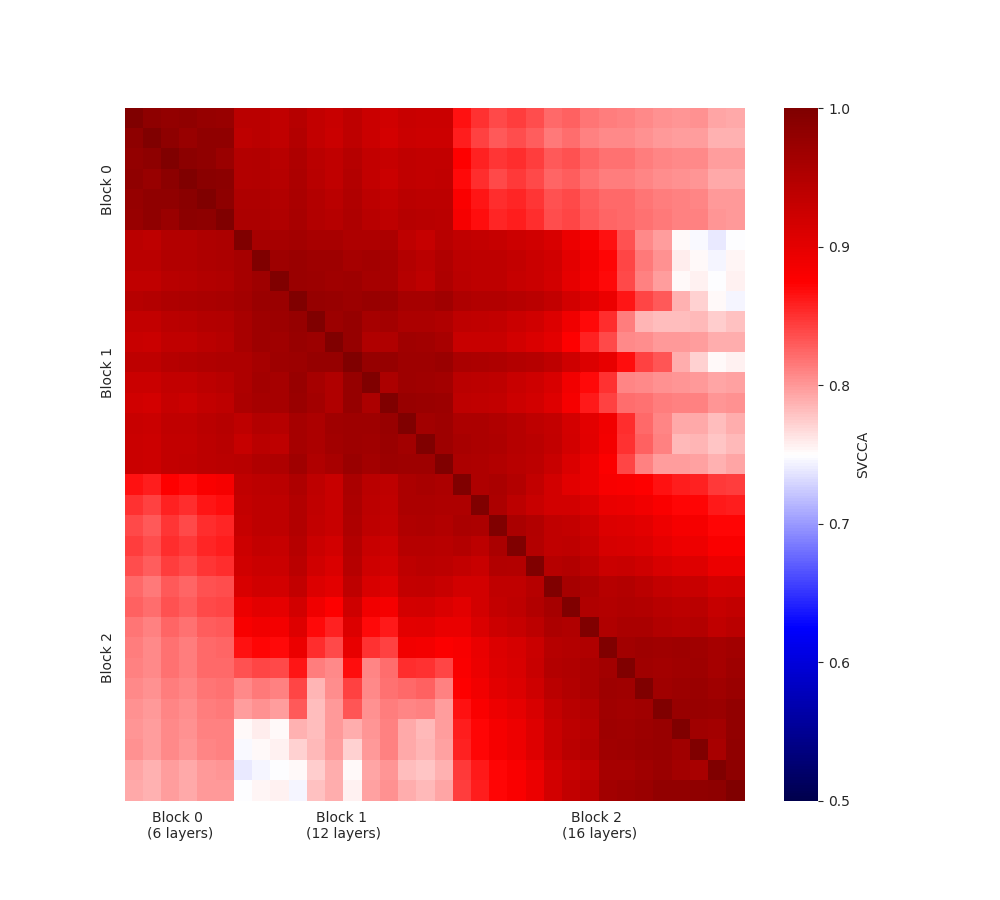}%
            \label{fig:tiny_densenet_svcca}%
        }%
        }
        \caption{Internal representation of DenseNet and its tiny version. The SVCCA is computed between each pair of layers. Networks are trained on Dx.}
        \label{fig:svcca_densenet_dx}
    \end{figure*}

    \textbf{Analysis of DenseNet121}: as we wanted to give a tiny version of DenseNet (121 layers and 11M parameters), we analyzed its internal representation on Dx problem. In order to analyze the representation learnt inside this network, we computed the Singular Vector Canonical Correlation Analysis (SVCCA) \cite{raghu2017svcca} between the outputs of all pairs of layer inside every block. Formally, we define a set of neurons $\{\mathbf{z}_i^l\}_{i\in[1..hwcd]}$ for each layer $l$ where $(c, h, w, d)$ represent the number of channels, height, width and depth of the feature maps of layer $l$ respectively; and $\mathbf{z}_i^l = (\mathbf{z}_i^l(x_1), ...,\mathbf{z}_i^l(x_N))\in \mathbb{R}^N$ is the response of neuron $i$ to the entire test set (of size $N$). In this way, we can compute the CCA between 2 blocks of data $\{\mathbf{z}_i^{l_1}\}_{i\in[1..h_1w_1c_1d_1]}$ and $\{\mathbf{z}_i^{l_2}\}_{i\in[1..h_2w_2c_2d_2]}$ for 2 layers $l_1$ and $l_2$ since all vectors lie in the same space $\mathbb{R}^N$ (we also computed a Singular Value Decomposition (SVD) before the computation of the CCA to remove the noisy neurons, as described in \cite{raghu2017svcca}). We chose to keep only 50\% of the explained variance since $N \ll hwcd$ in our experiments ($N=394$ and $hwcd > 10^4$)  and we observed that a lot of neurons were noisy. Results are plotted in figure \ref{fig:densenet121_svcca}. 
            
    \textbf{Tiny-DenseNet}: we first observed that the blocks 1 and 2 (starting from 0) of DenseNet121 were highly correlated, which suggested a redundancy.  In particular, it suggested that the features learnt inside the \nth{3} block were just copied from the second block and the specialization of the network to the prediction task did not occur in block 2. It was then natural to remove the block 2 from DenseNet121, assuming that the receptive field of a neuron before the FC layer would remain big enough for the 3 clinical tasks (its size is $32\times32\times32$ for a an input size $128\times128\times128$ with DenseNet121 and it is halved when we remove the \nth{3} block). Also, we halved the growth rate from $k=32$ to $k=16$ and we called the resulting network \textit{tiny-DenseNet}, as it is $10\times$ smaller than DenseNet. As before, we plotted the SVCCA between the internal layer outputs of tiny-DenseNet in figure \ref{fig:tiny_densenet_svcca} and we noticed that, differently from DenseNet121 in figure \ref{fig:densenet121_svcca}, the strong correlation between blocks disappeared.
            
\subsection*{Convergence speed}
        \begin{figure*}[h!]
            \centering
            \includegraphics[width=\linewidth]{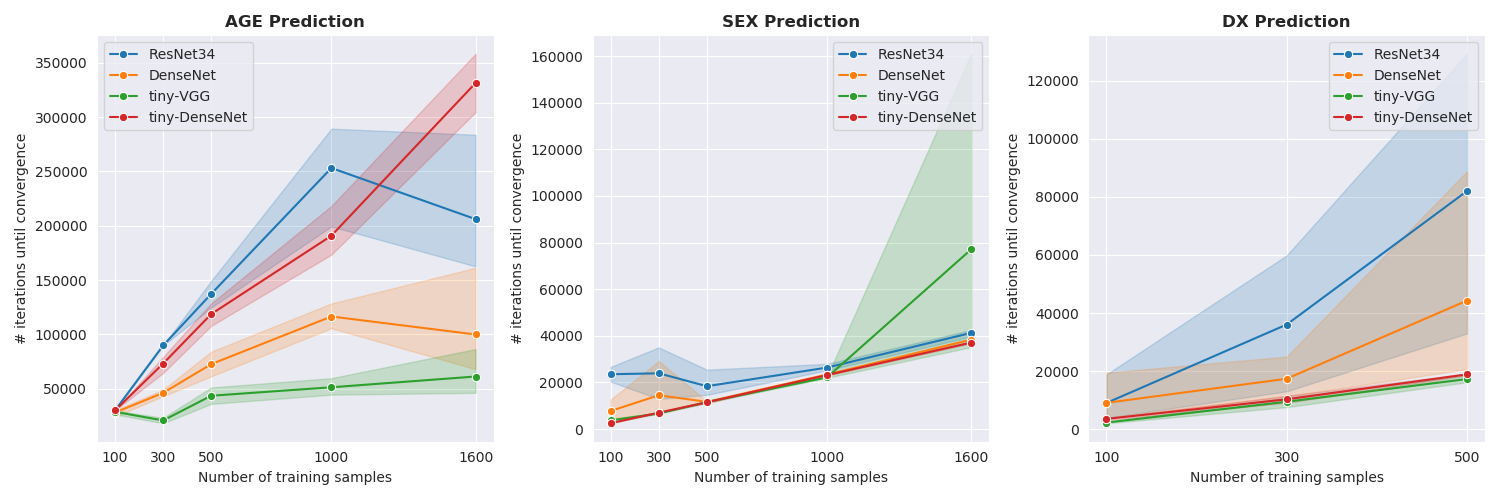}
            \includegraphics[width=\linewidth]{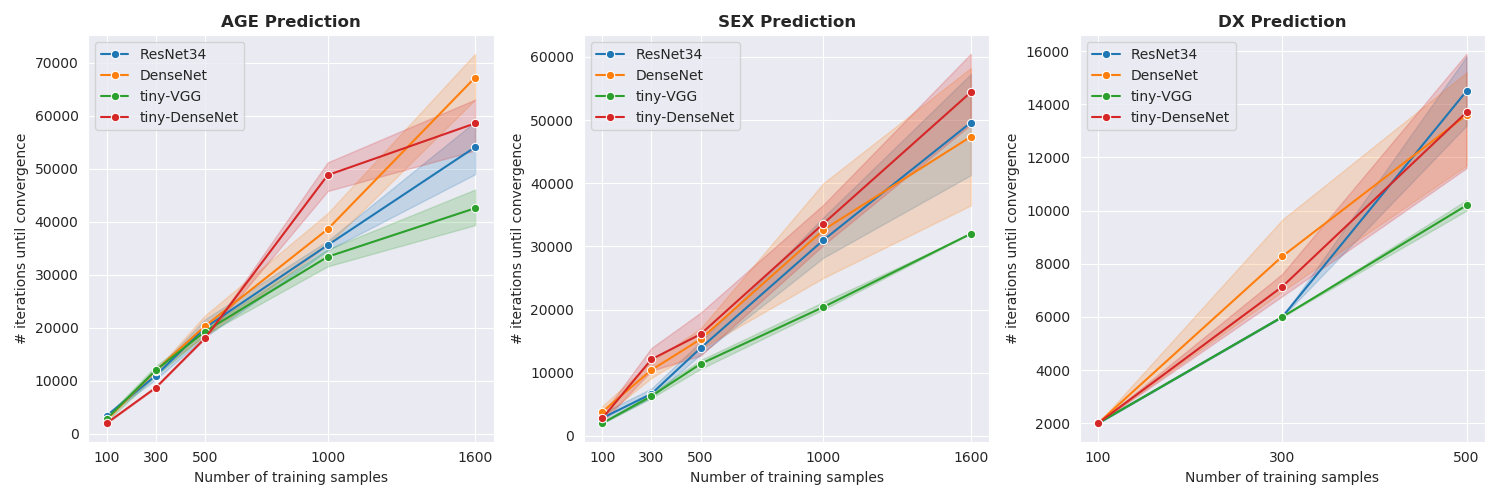}
            \caption{Convergence speed of the four main CNN architectures as we vary the number of training samples with i) VBM data (top) and ii) quasi-raw (bottom). The stopping criterion is defined through the variance of a rolling window on the validation loss. The convergence is reached when this loss remains stable for the next $k=20$ epochs (see section \ref{meth_learning_curves_convrgence} for more details).}
            \label{fig:convergence_speed}
        \end{figure*}
        
        We observe in figure \ref{fig:convergence_speed} that tiny-VGG constantly converges faster than any other networks for both quasi-raw and VBM data. Tiny-DenseNet also converges at the same rate for the 2 classification tasks (Dx and sex prediction). This is somewhat expected since they are the lightest networks among the ones tested (about $10\times$ less parameters than DenseNet and $60\times$ less than ResNet34). \\
        Surprisingly, on quasi-raw data, all the models converge i) faster than on VBM data (about $5\times$ faster on age prediction) and ii) at the same rate (ResNet34 is the slowest globally on VBM data while it performs similarly with tiny-DenseNet on quasi-raw data).

\subsection*{2D slice-level approach}
    
    \begin{table*}[h!]
        \centering
        \resizebox{\textwidth}{!}{
        \begin{tabular}{|c|c||cc|cc||cc|cc|}
        \hline
        \multirow{3}{*}{Preprocessing} & \multirow{3}{*}{Architecture} & \multicolumn{4}{c||}{\textbf{Sex}} & \multicolumn{4}{c|}{\textbf{Dx}} \\
         &  &  \multicolumn{2}{c}{BAcc $\uparrow$} & \multicolumn{2}{c||}{AUC $\uparrow$} & \multicolumn{2}{c}{BAcc $\uparrow$} & \multicolumn{2}{c|}{AUC $\uparrow$} \\
        \cline{3-10}
         &  &  3D & 2D & 3D & 2D & 3D & 2D & 3D & 2D \\
        \hline
         \multirow{3}{*}{VBM} & ResNet18 & $\mathbf{0.83 \pm 0.01}$ & $\mathbf{0.78 \pm 0.03}$ & $\mathbf{0.91\pm 0.01}$ & $\mathbf{0.88 \pm 0.03}$ & $0.71 \pm 0.01$ & $0.56\pm 0.04$ & $0.78\pm 0.01$ & $0.64\pm 0.01$ \\
         & VGG11 & $0.72 \pm 0.015$ & $0.75 \pm 0.03$ & $0.80\pm 0.02$ & $0.85 \pm 0.02$ & $0.66\pm 0.03$ & $\mathbf{0.63\pm 0.01}$ & $0.71\pm 0.009$ & $\mathbf{0.68\pm 0.01}$ \\
          & DenseNet121 & $0.83\pm 0.01$ & $0.68 \pm 0.03$ & $0.91\pm 0.01$ & $0.81 \pm 0.03$ & $\mathbf{0.72\pm 0.02}$ & $0.58\pm 0.04$ & $\mathbf{0.78\pm 0.01}$ & $0.67\pm 0.02$ \\
         \hline\hline
         \cline{2-10}
         \multirow{3}{*}{Quasi-Raw} & ResNet18 & $0.63\pm 0.02$ & $0.60 \pm 0,05$ & $\mathbf{0.84\pm 0.01}$ & $\mathbf{0.72 \pm 0.04}$ & $0.63\pm 0.02$ & $0.54 \pm 0.02$ & $0.66\pm 0.01$ & $0.59 \pm 0.02$ \\
         & VGG11 & $0.54\pm 0.02$ & $0.60 \pm 0.08$ & $0.68\pm 0.05$ & $0.72 \pm 0.05$ & $0.53\pm 0.04$ & $\mathbf{0.61 \pm 0.03}$ & $0.61\pm 0.02$ & $\mathbf{0.68 \pm 0.02}$ \\
         & DenseNet121 & $\mathbf{0.68\pm 0.05}$ & $\mathbf{0.62 \pm 0.08}$ & $0.81\pm 0.01$ & $0.72 \pm 0.05$ & $\mathbf{0.65\pm 0.02}$ & $0.56 \pm 0.04$ & $\mathbf{0.72\pm 0.01}$ & $0.62 \pm 0.04$ \\
         \hline
        \end{tabular}}
        \caption{Comparison between 2D and 3D models in the small data regime ($N=500$ training samples). The models are tested on the independent test set BSNIP and the same 5-fold RLT CV has been used as in section \ref{cross_val_strategy}. Even if 3D models have more parameters, they successfully capture the 3D brain anatomical structure by outperforming consistently their 2D counterpart. VGG11 is the only exception with lower performance in 3D, which can be due to its last 3 Fully-Connected layers that heavily increases the model size ($\ge 50$M in 3D). Best results for 2D and 3D CNN are reported in bold. BAcc=Balanced Accuracy, AUC=Area Under ROC Curve}
        \label{tab:2d_cnn_perf_N500}
    \end{table*}
    
    We also compared the performance of a 2D approach by decomposing each MRI scan into chunks of 3 consecutive axial slices. These slices are given to a 2D CNN with 3 input channels and we employed the same 5-fold RLT CV strategy as described in section \ref{cross_val_strategy}. We only used $N=500$ training samples and we reported the results on BSNIP, making them directly comparable with table \ref{tab:overall_results_sex_age_dx_archi_N500}. The predictions are performed at the subject-level by taking the median of the individual slice prediction and we reported the results in table \ref{tab:2d_cnn_perf_N500}. 

\subsection*{Data Augmentation Strategies}

        \begin{table*}[h!]
                 \centering
                 \setlength\tabcolsep{2pt}
                 \begin{tabular}{|>{\centering}m{0.15\textwidth}|>{\centering}m{0.2\textwidth}|m{0.5\textwidth}|>{\centering\arraybackslash}p{0.15\textwidth}|}
                 \hline
                     \textbf{Application} & \textbf{Transformation} & \multicolumn{1}{|c|}{\textbf{Details}} & \textbf{Hyperparameters}\\
                     \hline
                     \multirow{5}{*}{\vspace{-1cm}Computer Vision} & Flip & The images are flipped randomly along the 3 directions (axial, sagittal, coronal). & \xmark \\
                     \cline{2-3}
                     & Gaussian Blur & A Gaussian filter is applied to input images with a full width at half maximum (FWHM) uniformly sampled in $[\alpha, \beta]$ & $\text{FWHM} \in [0.35mm, 3.5mm]$ \\
                     \cline{2-3}
                     & Gaussian Noise & A Gaussian noise is added with a variance $\sigma$ uniformly sampled in $[\alpha, \beta]$. & $\sigma \in [0.1, 1]$ \\
                     \cline{2-3}
                     & Random Crop (+Resize) & The images are cropped at a random location, reducing the input shape by $p$\% in every direction, and resized linearly to match the input size. & Patch $p=70\%$ \\
                     \cline{2-3}
                     & Affine & The images are randomly translated up to $k$ voxels in every direction and rotated up to $\alpha$ degrees. & $k=10$ voxels, $\alpha=5^{\circ}$\\
                     \hline
                     \multirow{5}{*}{\vspace{-5cm}Neuroimaging} & k-space Ghosting Artefact \cite{zhuo2006_MRI_artefacts} & $n$ lines in the k-space are randomly distorted to mimic the errors that may happen during the k-space line inversion step in an echo-planar imaging acquisition. & $n=10$ \\
                     \cline{2-3}
                     & k-space Motion Artefact \cite{shaw2019_motion_artefact} & The image is successively randomly linearly transformed ($n_{sim} \times$, up to $\alpha^{\circ}$ rotation, $t$ voxels translation) to reproduce the head motion artefact observed during an acquisition. The 3D Fourier transforms of these images are then combined to form a single k-space, which is transformed back to the original space. & $n_{sim}=3$, $\alpha=40^{\circ}$, $t=10$ voxels\\
                     \cline{2-3}
                     & k-space Spike Artefact \cite{zhuo2006_MRI_artefacts} & $n$ points with very high or low intensity are added randomly in the k-space reproducing the bad data points obtained with gradients applied at a very high duty cycle. It results in dark stripes in the original image. & $n=10$\\
                     \cline{2-3} 
                     & Bias-Field Artefact \cite{van1999} & The voxel intensities are modulated by a polynomial function (order 3, coeff. magnitude $m$) whose coefficients are randomly sampled. It models the artefacts in the low-frequency range produced by the inhomogeneity of the static magnetic field inside the MRI scanner. & $m\in [-0.7, 0.7]$ \\
                     \cline{2-3}
                     & Swap \cite{chen2019self_supervision}  & $n$ pairs of patches with shape $15\times 15\times 15$ are randomly swapped. Originally created as a self-supervision task to learn meaningful semantic features, the network is expected to use the context around each patch in order to find its original location and internally reconstruct the image. & $n=20$\\
                     \hline
                 \end{tabular}
                 \caption{Description of the data augmentation strategies considered in our experiments. The input image always correspond to the pre-processed MR image. All the k-space artefacts have been implemented in the Python library TorchIO \cite{perez2020torchio}. }
                 \label{tab:data_augmentation_description}
        \end{table*}
    
    All the detailed hyper-parameters used for the experiments with Data Augmentation along with their description can be found in table \ref{tab:data_augmentation_description}.

\subsection*{MC-Dropout}
    \begin{figure*}[h!]
        \centering
        \includegraphics[width=\linewidth]{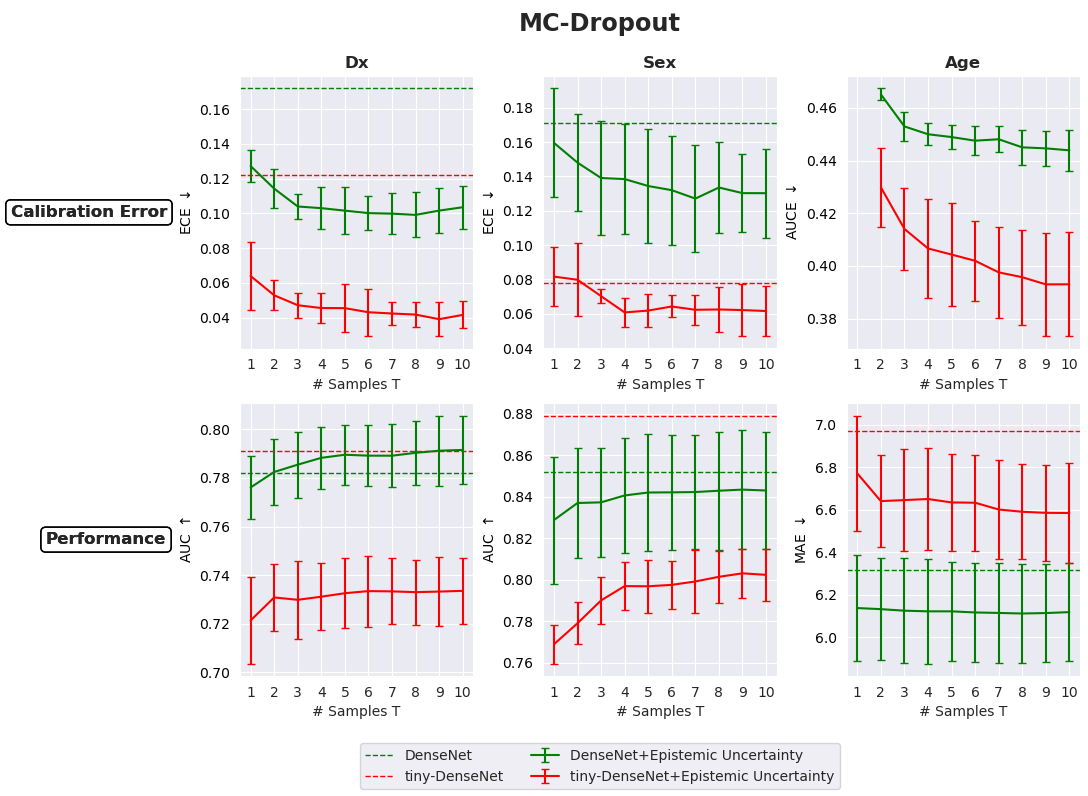}
        \caption{Performance of big and small networks (resp. DenseNet and tiny-DenseNet) as we increase the number of feed-forward passes $T$ when $N=500$ training samples. We integrated Concrete Dropout \cite{gal2017concrete_dropout} inside the CNN architecture to model the epistemic uncertainty. The dashed lines represent the baselines of the deterministic models (without Concrete Dropout).}
        \label{fig:MC_dropout}
    \end{figure*}
    
    MC-Dropout has been introduced in \cite{gal2016phd, gal2016dropout} as a simple way to capture epistemic uncertainty by putting a Bernouilli prior $\mathcal{B}(p)$ on the model's weight $w$ (the resulting network is referred to as a Bayesian Neural Network, see for instance \cite{kendall2017uncertainties}). In practice, adding dropout inside a network is not new \cite{srivastava2014dropout} but it was mainly applied as a regularization technique to limit over-fitting. In the Bayesian context, dropout is applied both at training time and test time. Specifically, a single feed-forward pass at test time corresponds to a sampling $\hat{w}\sim p(w)$ and to compute $f_{\hat{w}}(x)$ for a given input image $x$. Averaging these outputs for several $\hat{w}\sim p(w)$ gives an approximation of the posterior $p(y|x)$ in the classification case. \\
    One main drawback of MC-Dropout is the tuning of the dropout hyper-parameter $p$ by grid-search. One way to avoid this computationally expensive grid-search is to learn this hyper-parameter automatically during training, a technique referred to as Concrete Dropout \cite{gal2017concrete_dropout}. We used this technique in this paper.

\subsection*{Calibration Metrics}

\subsubsection*{Calibration for classification}
    Let's assume that a DNN outputs a class prediction $y$ as well as a confidence estimate $\hat{p}$ (usually the maximum probability after softmax) for a given $x$. We want to evaluate this estimation of confidence through a "calibration curve". Intuitively, if a network outputs a class $y=0$ with a confidence level $\hat{p}=0.6$, then we would like that over 100 predictions of samples belonging to class 0, 60 are correctly classified. More formally, we introduce a notion of accuracy for a given confidence level $p$ as $p(\mathbf{y}=y|\mathbf{\hat{p}}=p)$. A perfectly calibrated model should always verify:
    \begin{equation*}
        \forall p\in [0,1], \forall y \in [1..K], p(\mathbf{y}=y|\mathbf{\hat{p}}=p)=p
    \end{equation*} 
    in a classification problem with $K$ classes. In practice, this accuracy has to be estimated for various confidence levels $p$ and given a class $k$. To do so, we discretize uniformly the predicted confidence levels $\hat{p}=(\hat{p}_i)$ into $L$ bins $I_l=[\frac{l-1}{L}, \frac{l}{L})$ and compute the accuracy of the predictions over each bin $\hat{P}_l = \{i | \frac{l-1}{L} \le \hat{p}_i < \frac{l}{L}\}$ by:
    \begin{equation*}
        acc(\hat{P}_l)= \frac{1}{|\hat{P}_l|}\sum_{i\in \hat{P}_l} \mathbb{1}_{y_i=k}
    \end{equation*}
    
    The estimation of the confidence level associated to the bin $l$, independent from class $k$, is then:
    \begin{equation*}
        conf(\hat{P}_l) =  \frac{1}{|\hat{P}_l|}\sum_{i\in \hat{P}_l} \hat{p}_i
    \end{equation*}
            
    In a perfectly calibrated model, we expect $\forall l\in [1..L], acc(\hat{P}_l)=conf(\hat{P}_l)$. One visual way to check the model calibration is to plot the accuracy function of confidence, the ideal case being $acc=conf$. A usual statistic derived from this calibration curve is called Expected Calibration Error (ECE) and it is defined as \cite{guo2017calibration}:
    \begin{equation*}
        ECE = \sum_{l=1}^L\frac{|\hat{P}_l|}{n}\left(acc(\hat{P}_l) - conf(\hat{P}_l)\right)
    \end{equation*}
    where $n$ is the total number of samples. We systematically used this metric to measure calibration on sex prediction and Dx.
    
\subsubsection*{Calibration for regression}

    We can extend the ECE metric to the regression case, as detailed in \cite{gustafsson2020bench}. Briefly, assuming that the model outputs a mean $\mu$ and variance $\sigma^2$ of a Gaussian distribution for a given $x$, we can build a confidence interval $CI(p)=[\mu - \Phi^{-1}\left(\frac{p+1}{2}\right)\sigma, \mu + \Phi^{-1}\left(\frac{p+1}{2}\right)\sigma]$ associated to a confidence level $p$ (where $\Phi$ is the Cumulative Distribution Function, CDF, of $\mathcal{N}(0, 1)$). We can compute the proportion $\hat{p}$ of true target points $y\in \mathbb{R}$ that lie in $CI(p)$, for all $p\in [0, 1]$. From this, similarly to ECE, we can deduce the Area Under the Calibration Error (AUCE) of $|\hat{p} - p|$. 

\subsection*{Convergence curves at $N=500$}
    \begin{figure*}
        \centering
        \includegraphics[width=\textwidth]{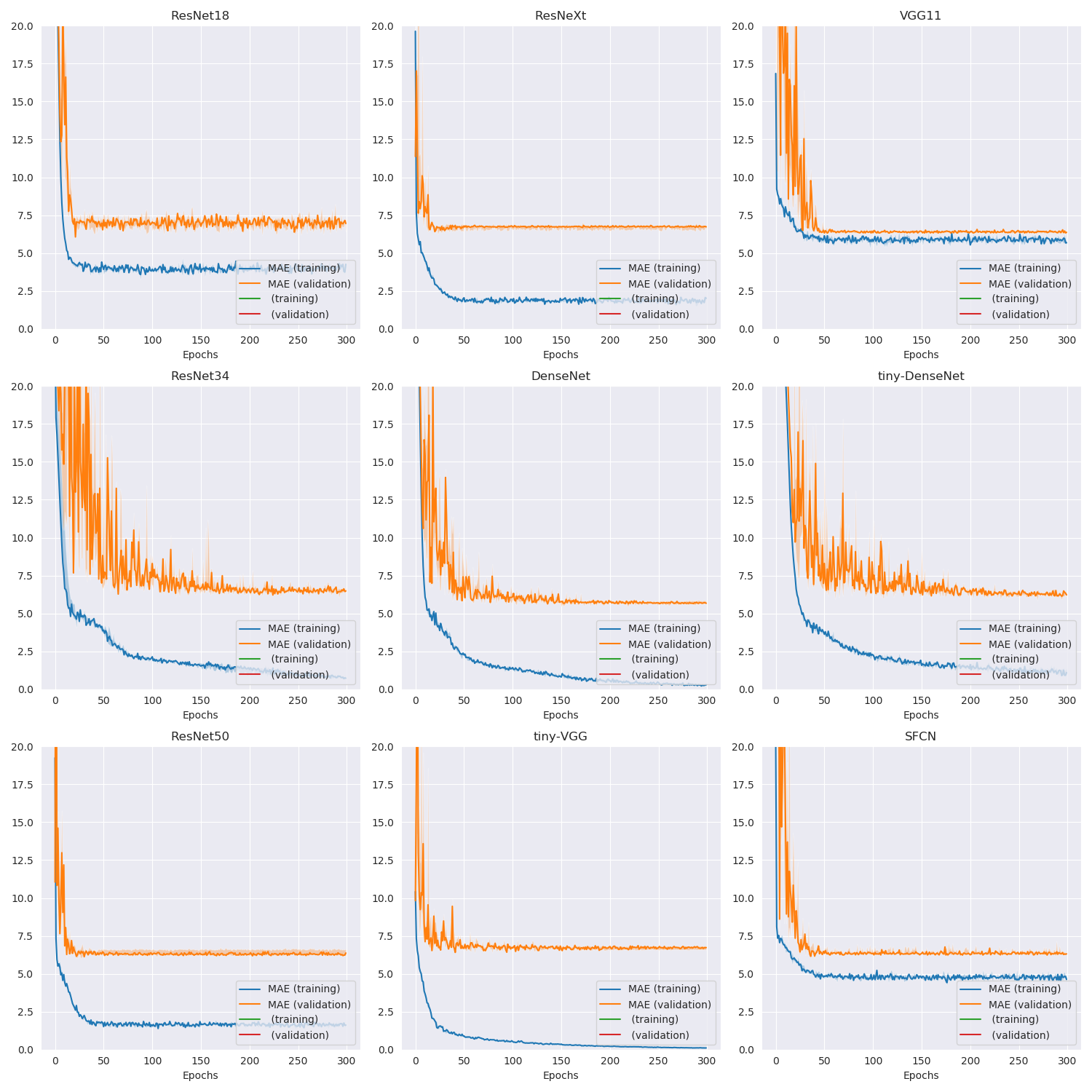}
        \caption{Training and validation losses for \textbf{age prediction} using only $N=500$ training samples and VBM data. The losses correspond to the $\ell_1$ error between the true age and the predicted age at each epoch for all CNN.}
        \label{fig:age_N_500_cnn_losses}
    \end{figure*}
    \begin{figure*}
        \centering
        \includegraphics[width=\textwidth]{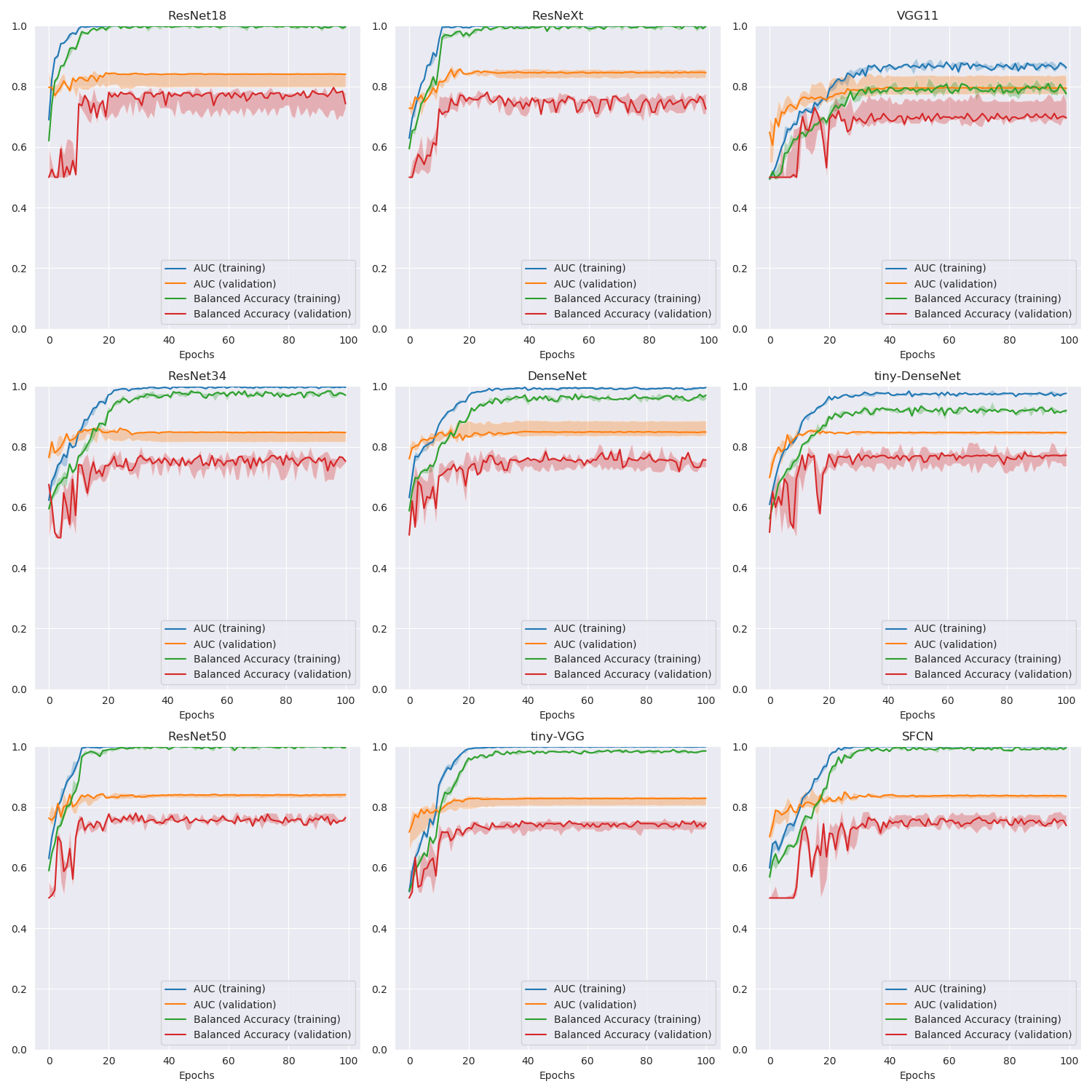}
        \caption{ROC-AUC and Balanced Accuracy for \textbf{Dx} with $N=500$ training samples and VBM data. }
        \label{fig:dx_N_500_cnn_losses}
    \end{figure*}
    
    We reported the convergence curves of all tested models on age regression and Dx figures \ref{fig:age_N_500_cnn_losses} and \ref{fig:dx_N_500_cnn_losses} with $N_{train}=500$.  These plots motivate the stopping criterion used throughout our experiments and described section \ref{meth_learning_curves_convrgence}. In particular, we see a clear plateau each time for every model without any evolution in performance given a long enough training.

\end{document}